\documentclass{iopjournal}
\usepackage{multicol}
\usepackage{graphicx}
\usepackage[caption=false]{subfig}
\usepackage{amsmath} 

\begin{document}

\articletype{PAPER} 

\title{Length Control of Pneumatic Artificial Muscles Inspired by Intrafusal--Extrafusal Muscle Interactions}

\author{Ko Sakamoto$^{1}$\orcid{0009-0009-2238-4977}, Yoichi Masuda$^1,*$\orcid{0000-0001-7365-5022}, Sei-ichiro Takeda$^1$\orcid{0009-0002-4139-6104}, Masato Ishikawa$^1$}

\affil{$^1$Department of Mechanical Engineering, The University of Osaka, Osaka, Japan}

\affil{$^*$Author to whom any correspondence should be addressed.}

\email{masuda@mech.eng.osaka-u.ac.jp}


\begin{abstract}
Pneumatic artificial muscles (PAMs) are widely used in pneumatic musculoskeletal robots due to their lightweight nature and high power output. 
PAM control also typically relies on electro--pneumatic valves, electronic sensors, and peripheral circuits such as A/D converters, which increases system weight, wiring complexity, and the risk of damage to electronic components.
To overcome these challenges, inspired by the intrafusal--extrafusal muscle interactions in biological systems, this paper proposes a length control method for PAMs in which the pneumatic feedback loop is implemented mechanically, while the target length is commanded by a minimal electronic actuator.
We first examined flow path conditions to suppress oscillations caused by pneumatic delays.
Experiments demonstrated a tracking error of $\le2.01\%$ for input frequencies $\le0.20\,\mathrm{Hz}$.
\end{abstract}

\section{Introduction}
    To date, many studies have investigated pneumatic musculoskeletal robots that feature compliance and multiple degrees of freedom \cite{hosoda2006biped,niiyama2007,yamada2011neural,lei2016dynamic}. 
    Pneumatic artificial muscles (PAMs) offer high flexibility and high power output despite their light weight. 
    On the other hand, the strong nonlinearity originating from the material properties and structural characteristics of PAMs, as well as air compressibility, makes modeling challenging and leads to increased complexity in sensing systems and control laws. 
    Therefore, in this study, we discuss a simple method to control the length of PAMs. 

    Various approaches have been proposed for the length sensing and control of PAMs. 
    Tondu et al. \cite{tondu2000modeling} developed a model considering the physical and geometric characteristics of McKibben PAMs and proposed a control method for an antagonistic drive system using sliding mode control with pressure sensors. 
    Furthermore, Xing et al. \cite{Xing2010observer} proposed a tracking control method based on disturbance estimation using a nonlinear disturbance observer. 
    Yokoyama et al. \cite{yokoyama2017mckibben} proposed a position control method for PAMs based on contraction ratio estimation using a Kalman filter.    
    Sensors for measuring the state of pneumatic artificial muscles have also been actively developed. 
    Hitzmann et al. \cite{hitzmann2019muscle} developed a sensor that estimates the internal pressure and length changes of PAMs by measuring the electrical resistance variation of a conductive fabric attached to the muscle. 
    Takahashi et al. \cite{takahashi2023reflex} improved the responsiveness of a robot by utilizing the sensor developed by Hitzmann et al. 
    Similarly, Park et al. \cite{park2013smart} developed a flexible sensor utilizing micro-channels filled with a conductive liquid and proposed a method to detect geometric changes based on variations in electrical resistance. 
    Felt et al. \cite{felt2016contraction} developed a PAM integrated with a sensor by embedding conductive wires into the sleeve covering a McKibben PAM. 
    In general, electronic pneumatic control systems require electro--pneumatic valves, electronic sensors, and peripheral circuits such as A/D converters, which increases system weight, wiring complexity, and the risk of damage to electronic components.
    To address this issue, this paper discusses a length control method for PAMs that reduces the need for electronic sensing and electro--pneumatic regulation by replacing the pneumatic feedback loop with a mechanically implemented one.

    Mechanically implemented fluid control has a long history \cite{bennett2001review}.
    From the perspective of position and force control in linear actuator, Turnbull \cite{turnbull1959hydraulic} discussed the responsiveness of a hydraulic servomechanism that uses a position command as the input and the actuator position as the output. 
    Moriguchi et al. \cite{moriguchi2018pneumatic} investigated the characteristics of a position control system using a pneumatic cylinder and a cylinder positioner. 
    However, servo valves and positioners for fluid flow control are not suitable for controlling lightweight and flexible PAMs, as they are generally rigid and often assume the use of double-acting air cylinders. 

    In recent years, compact and lightweight valves that enable continuous-valued fluid control via mechanical feedback have been developed in electronics-free robotics \cite{wang2025review}.
    As a recent example of such continuous control, Choe et al. \cite{Choe2023SoftValve} developed a soft valve that continuously varies flow resistance in response to tensile strain and proposed a method for proportional control of pneumatic actuators.
    Furthermore, Stanley et al. \cite{Stanley2024FluidicTransistor} developed a four-terminal fluidic transistor suitable for high-speed switching and realized a closed-loop position control system.
    From the perspective of bio-inspired control of artificial muscles, an electronics-free device inspired by Golgi tendon organs (mechanoreceptors) has achieved continuous pneumatic pressure output corresponding to muscle force \cite{masuda2022forcereceptor}.
    However, continuous muscle length control remains a challenge. 

    In this study, we focus on the control mechanism of animals to realize continuous length control of PAMs. 
    The receptors embedded in animal muscles measure physical information within the muscle and configure sensory--motor feedback via motor neurons in the spinal cord. 
    The proprioceptor involved in muscle length control is called the muscle spindle \cite{kandel2012principles}. 
    The muscle spindle has a distinctive structure; although it functions as a receptor (sensor), it contains small muscles (actuators) called the intrafusal muscles.
    Intrafusal muscles are innervated by efferent nerves called $\gamma$ motor neurons and are driven in response to commands from the higher-order neural system. 
    Importantly, mechanical interactions occur between intrafusal and extrafusal muscles (skeletal muscles), as the intrafusal muscles are arranged in parallel with the extrafusal muscles. 
    To understand and utilize the biological mechanism of muscle length control, it is necessary to consider the interactions between the intrafusal muscles contained within the muscle spindle and the extrafusal muscles. 

    This study proposes and evaluates the characteristics of a muscle length control method that combines a mechanically implemented pneumatic feedback with minimal electronic control, inspired by the interactions between intrafusal and extrafusal muscles. 
    In the field of neurophysiology, it has traditionally been explained that muscle spindles respond to muscle length and its rate of change (velocity) \cite{kandel2012principles}.
    However, recent research \cite{blum2017force} suggests the possibility that muscle spindles respond to muscle force. 
    Based on this finding, Masuda et al. \cite{masuda2025robomech} verified the principle of a length control device for PAMs, inspired by the possibility that muscle spindles respond to muscle force. 
    By extending this result, this paper reports on the development of an actuator system for performance evaluation, the investigation of flow path conditions effective in suppressing system oscillations, and the evaluation of the length control and load compensation performances of the PAMs.
    This actuator system receives length commands from a host computer via a daisy-chained servo motor. When the intrafusal wire displacement and the extrafusal PAM length remain synchronized, the PAM length can be geometrically estimated from the motor angle and pulley geometry.
    Furthermore, this system operates using only a single air supply line from a pneumatic source and wiring to the daisy-chained servo motor. 
    In this system, the pneumatic regulation loop is implemented by an electronics-free device without electronic circuits, while a commercially available servo motor is used only to prescribe the intrafusal target length. This configuration is expected to reduce the number of electronic sensing and regulation components, as well as wiring and maintenance costs, in musculoskeletal robots equipped with numerous PAMs.

\section{Intrafusal--Extrafusal Servo Actuator}
 \subsection{Overview of the Muscle Spindle and the Proposed System}
\begin{figure}[h]
          \centering
          \includegraphics[scale=0.275]{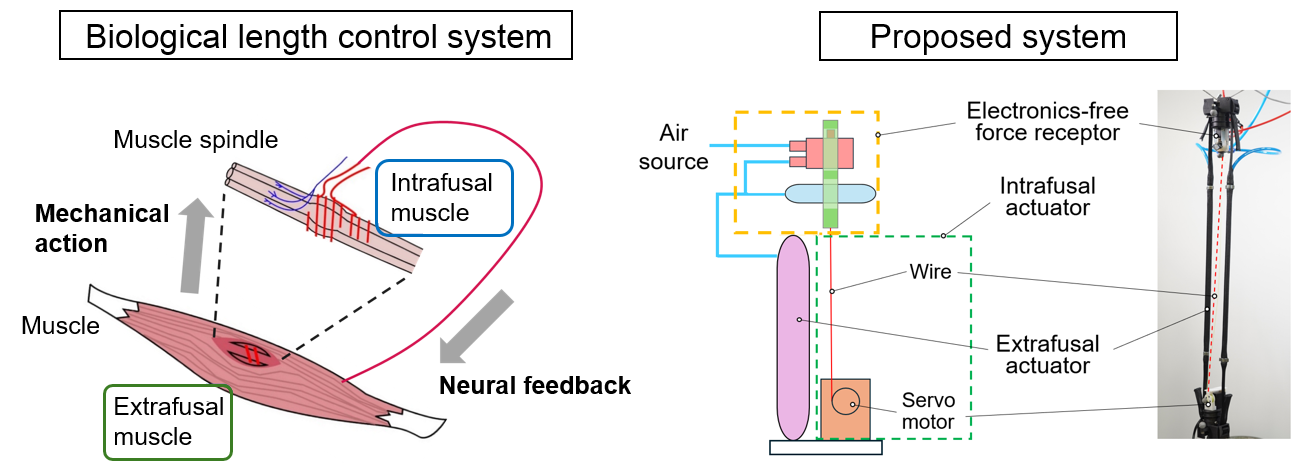}
          \caption{Overview of the biological length control system and the proposed InExSA.}
          \label{Fig:Structure}
       \end{figure}

        Fig.~\ref{Fig:Structure} illustrates the conceptual schematic of the biological muscle-muscle spindle system alongside the structure of the Intrafusal--Extrafusal Servo Actuator (InExSA) developed in this study. 
        In animals, muscle spindles are important proprioceptors involved in muscle length control \cite{kandel2012principles}.
        A muscle spindle consists of intrafusal muscle fibers and a capsule that covers them.
        Since both ends of the muscle spindle are connected to the extrafusal muscle (skeletal muscle), displacement and force are transmitted to the intrafusal muscle as the extrafusal muscle stretches. 

        The length control system of animal muscles operates under mechanical and neural interactions between intrafusal and extrafusal muscles. 
        When the extrafusal muscle stretches, its mechanical changes are transmitted to the intrafusal muscle. 
        Changes in the state of the intrafusal muscle are measured by a sensory region connected to the intrafusal muscle fibers and are transmitted to motor neurons in the spinal cord via afferent sensory nerves extending from the intrafusal muscle. 
        Upon receiving sensory signals, motor neurons output contraction commands to the extrafusal muscle based on their activity levels.
        In this paper, we discuss an actuator system that mimics the neuro--mechanical interactions between intrafusal and extrafusal muscles. 

        The question of what physical quantities are measured by the sensory regions of muscle spindles remains a subject of ongoing debate. 
        Although muscle spindles are traditionally described as encoding muscle length and velocity \cite{kandel2012principles}, a recent study suggests the possibility of force-related encoding \cite{blum2017force}.
        In our study, we control the length of the extrafusal actuator by using a mechanical device that converts the tension of the intrafusal actuator into air pressure. 

        In addition, we also use an electric servo motor (Dynamixel XC330-M288-T) to reproduce the contraction movement of the intrafusal muscle. 
        In animals, the intrafusal muscle is controlled by efferent $\gamma$ motor neurons and is driven in response to commands from the higher-order neural system \cite{kandel2012principles}. 

        This actuator system consists of an electronics-free force receptor (force-to-pneumatic pressure converter) that mimics the sensory region within a muscle spindle, an intrafusal actuator (motor-wire drive system), and an extrafusal actuator (McKibben PAMs). 
        The electronics-free force receptor outputs pneumatic pressure to the extrafusal actuator in accordance with the tension of the intrafusal actuator. 
        The intrafusal actuator is arranged in parallel with the extrafusal actuator and receives target length commands from a host computer. 
        In the following sections, we will introduce the operating principles of the individual components of this actuator, followed by an explanation of the operating principle of the entire system. 
        
    \subsection{Operating Principle of the Electronics-Free Force Receptor}
\begin{figure}[h]
          \centering
          \includegraphics[scale=0.14]{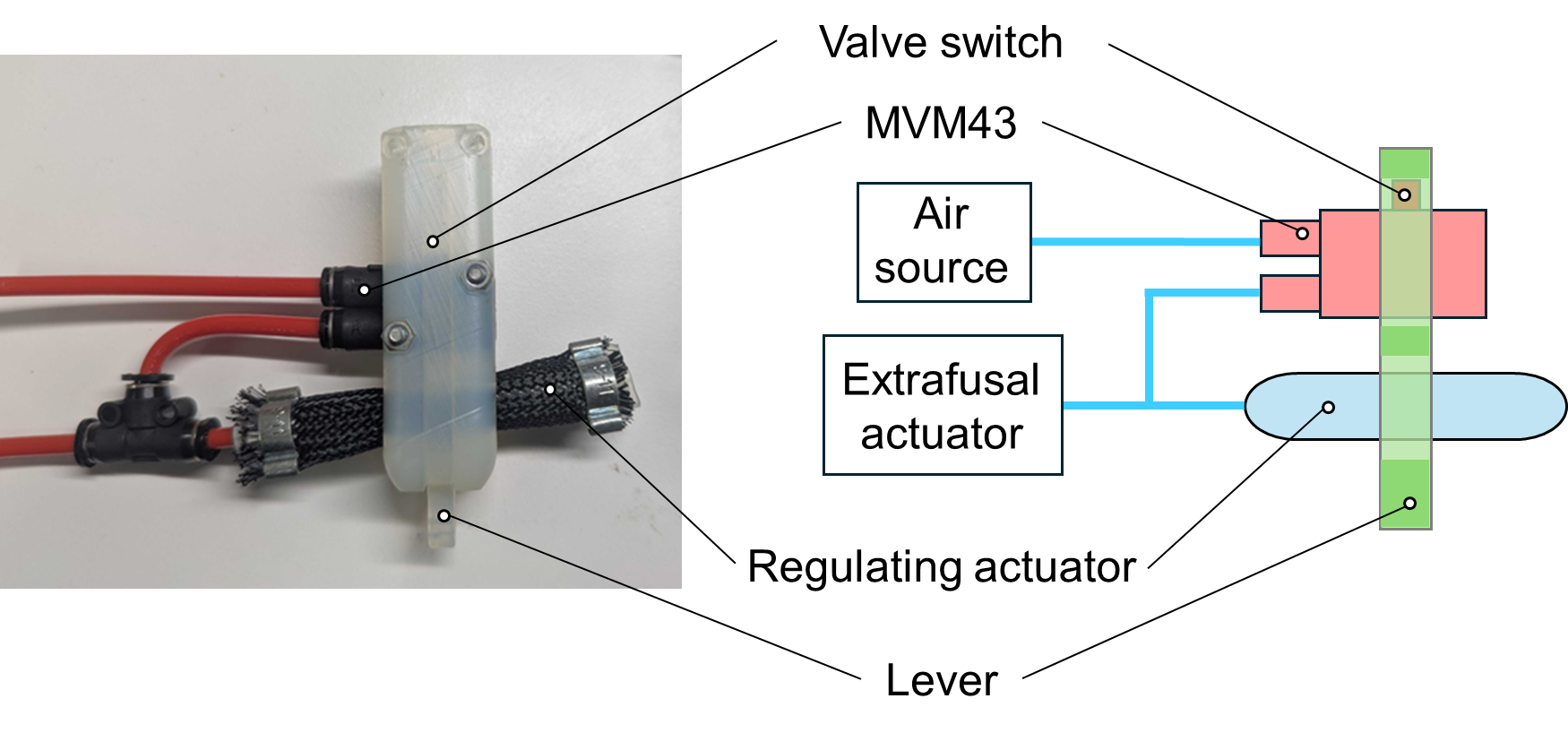}
          \caption{Electronics-free force receptor.}
          \label{Fig:Electronics-freeForceReceptor}
       \end{figure}

        In the proposed system shown in Fig.~\ref{Fig:Structure}, the electronics-free force receptor serves as the key component underlying the pneumatic feedback mechanism.
        The structure of the electronics-free force receptor, which continuously adjusts air pressure in response to external force, is shown in Fig.~\ref{Fig:Electronics-freeForceReceptor}. 
        This device consists of a normally-closed mechanical valve (PISCO MVM43, 3-port type), a lever connecting the valve switch to the external environment, and a regulating actuator (a small McKibben PAM). 

\begin{figure}[t!]
              \centering
              \includegraphics[scale=0.14]{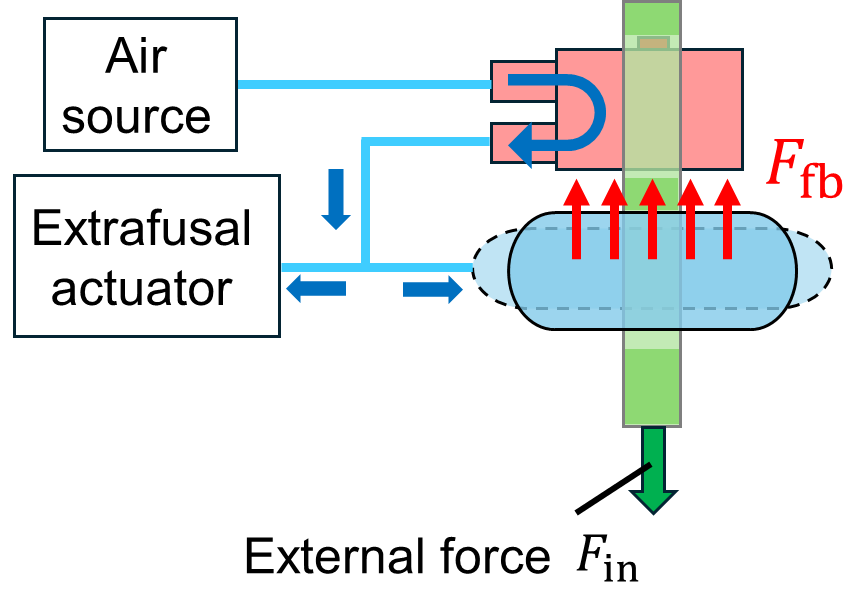}
              \caption{Schematic of the operational principle of electronics-free force receptor.}
              \label{Fig:Electronics-freeForceReceptor_Principle}
            \end{figure}

       Fig.~\ref{Fig:Electronics-freeForceReceptor_Principle} shows the operational principle of the electronics-free force receptor. 
       The basic operational principle of this device is based on previous research \cite{masuda2022forcereceptor}. 
       First, consider the case where no external force $F_{\mathrm{in}}$ is applied to the lever. At this time, since the mechanical valve is in a closed state, the air pressure supplied from the air source does not flow through the valve into the actuator. 
       Next, consider the case where an external force $F_{\mathrm{in}}$ is applied to the lever exceeding a certain threshold, and the mechanical valve is actuated. When the mechanical valve is opened, air flows into the regulating actuator and the extrafusal actuator, increasing the internal air pressure. 
       Then, the radial force $F_{\mathrm{fb}}(P(t))$ generated by the regulating actuator pushes up the lever, and simultaneously, the mechanical valve turns OFF. 
       Here, note that there is a slight deadband between ON and OFF of the valve switch, and the state remains in the deadband as both exhaust and intake are blocked. 
       Let $\overline{F}$ be the restoring force of the valve switch at this time. 
       Due to the above mechanism, the air pressure $P(t)$ inside the electronics-free force receptor is adjusted to satisfy the following equation. 
        
        \begin{equation}
          F_{\mathrm{in}} - F_{\mathrm{fb}}(P(t)) = \overline{F}. 
          \label{eq:Force-EquilibriumPoint}
        \end{equation} 

        Next, consider the case where $F_{\mathrm{in}}$ changes while the system is in the equilibrium state (\ref{eq:Force-EquilibriumPoint}). 
        When $F_{\mathrm{in}}$ increases, the valve switch that was pushed back by the regulating actuator turns ON again. 
        The valve increases the air pressure $P(t)$ inside the device, creating a new equilibrium state (\ref{eq:Force-EquilibriumPoint}). 
        Conversely, when $F_{\mathrm{in}}$ decreases, the valve switch turns OFF, decreasing the air pressure inside the device and creating a new equilibrium state (\ref{eq:Force-EquilibriumPoint}). 

    \subsection{Operating Principle of InExSA}
\begin{figure}[b]
              \centering
              \includegraphics[scale=0.2]{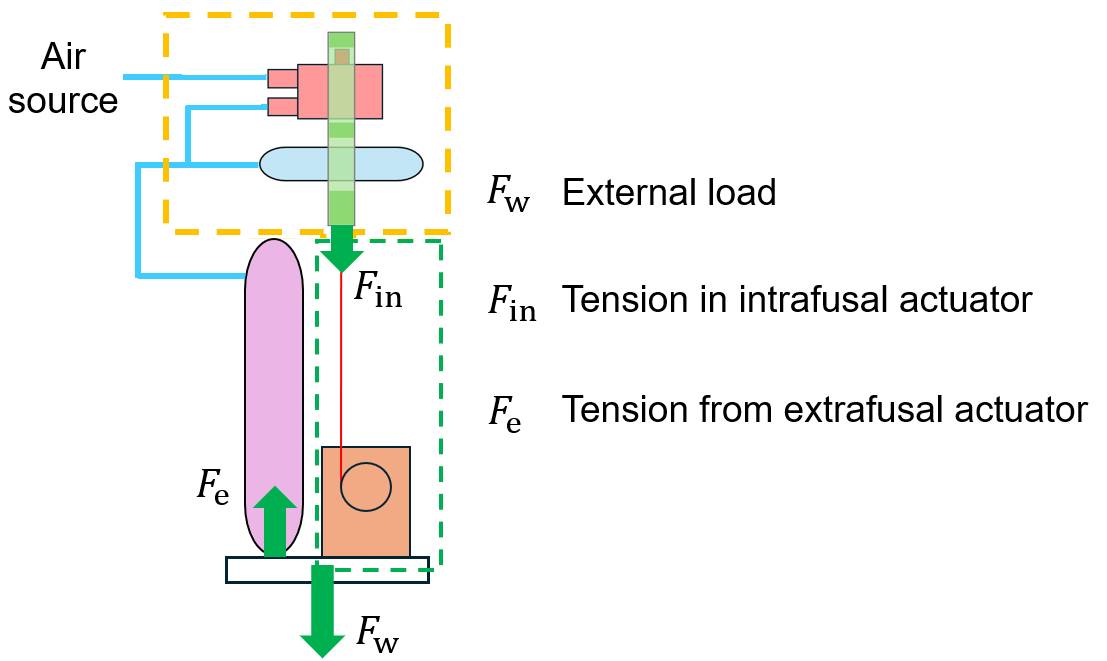}
              \caption{Schematic of the forces acting on the system.}
              \label{Fig:Force}
            \end{figure}

        Three primary forces act on this actuator system, as shown in Fig.~\ref{Fig:Force}. 
        Here, $F_{\mathrm{w}}$ is the external load, $F_{\mathrm{in}}(P(t))$ is the tension generated by the intrafusal actuator, and $F_{\mathrm{e}}(P(t))$ is the tension generated by the extrafusal actuator according to the air pressure $P(t)$. 

        First, consider winding the wire of the intrafusal actuator when the extrafusal actuator is in an extended state. 
        When an angular command in the contraction direction is applied to the electric motor, tension $F_{\mathrm{in}}$ is generated in the intrafusal actuator as the wire is wound up to the target length. 
        In this process, it should be noted that the maximum output force is set sufficiently lower than the load $F_{\mathrm{w}}$ to prevent damage to the electric motor.
        When the wire tension $F_{\mathrm{in}}$ exceeds the actuation force of the mechanical valve (turning the valve ON), compressed air is output from the electronics-free force receptor, which increases the tension $F_{\mathrm{e}}$ of the extrafusal actuator. 
        The tension $F_{\mathrm{e}}$ of the extrafusal actuator supports the load $F_{\mathrm{w}}$ and simultaneously reduces the tension $F_{\mathrm{in}}$ of the intrafusal actuator. 
        As $F_{\mathrm{in}}$ decreases, the mechanical valve switch begins to return to the OFF. 
        Upon reaching the deadband, the air inflow stops, satisfying the following equation. 
            \begin{equation}
              F_{\mathrm{in}}(P(t))+F_{\mathrm{e}}(P(t)) = F_{\mathrm{w}}.
              \label{eq:ControlLength}
            \end{equation}           
        It should be noted that $F_{\mathrm{in}}$ at this time satisfies (\ref{eq:Force-EquilibriumPoint}) within the electronics-free force receptor. 

        Conversely, consider the case where the wire of the intrafusal actuator is unwound in the extension direction while the extrafusal actuator is contracted by a certain amount. 
        By providing the electric motor with an angular command in the extension direction to increase the target wire length, the wire tension $F_{\mathrm{in}}$ temporarily decreases. 
        When the wire tension $F_{\mathrm{in}}$ decreases, the valve within the electronics-free force receptor turns OFF. 
        By exhausting the internal air from the system, the extrafusal actuator begins to extend, leading to a decrease in its tension $F_{\mathrm{e}}$.
        Once the extrafusal actuator extends to equal the target length (the wire length), the wire tension $F_{\mathrm{in}}$ increases again. 
        This wire tension $F_{\mathrm{in}}$ then pushes the mechanical valve switch; at the point where the valve reaches the deadband, the air outflow ceases, satisfying (\ref{eq:ControlLength}). 
        The above describes the operational principle of this actuator system during both contraction and extension. 

        Next, consider a case where the external load $F_{\mathrm{w}}$ increases while the extrafusal actuator is stationary at a constant length. 
        As $F_{\mathrm{w}}$ increases, both the tension of the extrafusal actuator $F_{\mathrm{e}}$ and the wire tension $F_{\mathrm{in}}$ increase simultaneously. 
        When $F_{\mathrm{in}}$ increases, the valve within the electronics-free force receptor turns ON, allowing compressed air to flow into the extrafusal actuator and thus increasing $F_{\mathrm{e}}$. 
        The increased tension $F_{\mathrm{e}}$ supports the load $F_{\mathrm{w}}$ while simultaneously reducing the tension $F_{\mathrm{in}}$ of the intrafusal actuator. 
        As $F_{\mathrm{in}}$ decreases, the mechanical valve turns OFF involving a deadband, which halts the air inflow and satisfies (\ref{eq:ControlLength}). 
        The above describes the principle by which the proposed actuator system compensates for external loads.         

\section{Performance Evaluation of the Proposed System}
    \subsection{Experimental Setup}
\begin{figure}[h]
              \centering
              \includegraphics[scale=0.2]{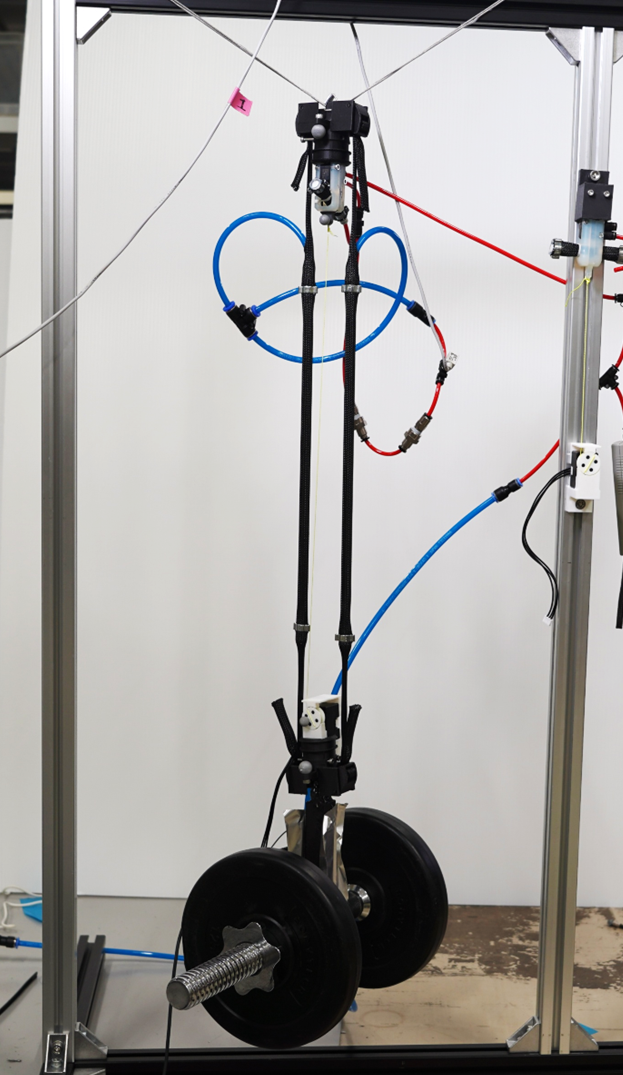}
              \caption{Experimental setup for performance evaluation.}
              \label{Fig:WeightLifter}
            \end{figure}

        Fig.~\ref{Fig:WeightLifter} shows the experimental setup developed in this study. 
        The specifications of this apparatus are presented in Table~\ref{tab:parameter}. 
        For the extrafusal actuator of this system, two McKibben PAMs with a contractile length of approximately $300\,\mathrm{mm}$ were used. 
        To avoid overloading the servo motor used as the intrafusal actuator, the torque limit was set so that the maximum tension generated in the wire was $25\,\mathrm{N}$. 
        A dumbbell was suspended from the end of the experimental apparatus as a weight to apply external force. 
        The load can be adjusted from $2.5\,\mathrm{kg}$ to $20.0\,\mathrm{kg}$ in increments of $2.5\,\mathrm{kg}$ by adding weight plates to the dumbbell. 
        Except for the analysis of load compensation performance in Section~\ref{LoadCompensationPerformance}, the weight of the dumbbell was fixed at $5.0\,\mathrm{kg}$ during the experiments. 
        The length of the extrafusal actuator was measured using a motion capture system (OptiTrack V120 Trio).
        Markers were fixed at two locations—the top and bottom of the apparatus—and the length of the extrafusal actuator was defined as the distance between these two points minus the length of the non-contractile parts. 
        The supply pressure from the pneumatic source was set to $0.75\,\mathrm{MPa}$. 
        
            \begin{table}[b]
                \caption{Experimental parameters}
                \begin{center}
                    \setlength{\tabcolsep}{3.5pt}
                    \begin{tabular}{c c}
                    \hline
                    \textbf{Parameter} & \textbf{value} \\
                    \hline
                    Number of extrafusal actuator & 2 \\
                    Natural length of extrafusal actuator & Approximately $300$~mm \\
                    External load & $5.0$~kg (except for  section \ref{LoadCompensationPerformance})\\
                    Wire tension limit (via torque limit) & $25$~N \\
                    Pneumatic supply pressure & $0.75$~MPa \\
                    \hline
                    \end{tabular}
                \label{tab:parameter}
                \end{center}
            \end{table}            

    \subsection{Oscillation Suppression in the Proposed System}
            Pneumatic actuation systems can exhibit oscillatory behavior due to delays and nonlinearities associated with air compressibility, piping, and valve dynamics.
            Therefore, in this paper, we adopted a passive stabilization approach by applying local flow restrictions to specific flow paths within the system. 
            Specifically, we identified the flow paths that contribute to oscillation suppression and investigated the orifice diameters required to achieve stable operation. 

            In the experiments, a square-wave target angle input ranging from $0\,\mathrm{deg}$ to $400\,\mathrm{deg}$ was applied to the servo motor. 
            Based on this input, the servo motor winds the wire of the intrafusal actuator (Fig.~\ref{Fig:Structure}) by $52$~mm at $1\,\mathrm{s}$ after the start of the experiment and unwinds it by $52$~mm at $11\,\mathrm{s}$ after the start of the experiment. 

            Furthermore, some one-way flow control valves (PISCO JKC4-01 series, constant-flow) with various orifice diameters were utilized to independently restrict and compare the flow rates in both the intake and exhaust directions of each flow path. 
            The inner diameter of each flow path prior to the application of flow restrictions is $2.5\,\mathrm{mm}$. 

            \subsubsection{Identification of Flow Restriction Paths Contributing to Oscillation Suppression}
\begin{figure}[h]
                  \centering
                  \includegraphics[scale=0.2]{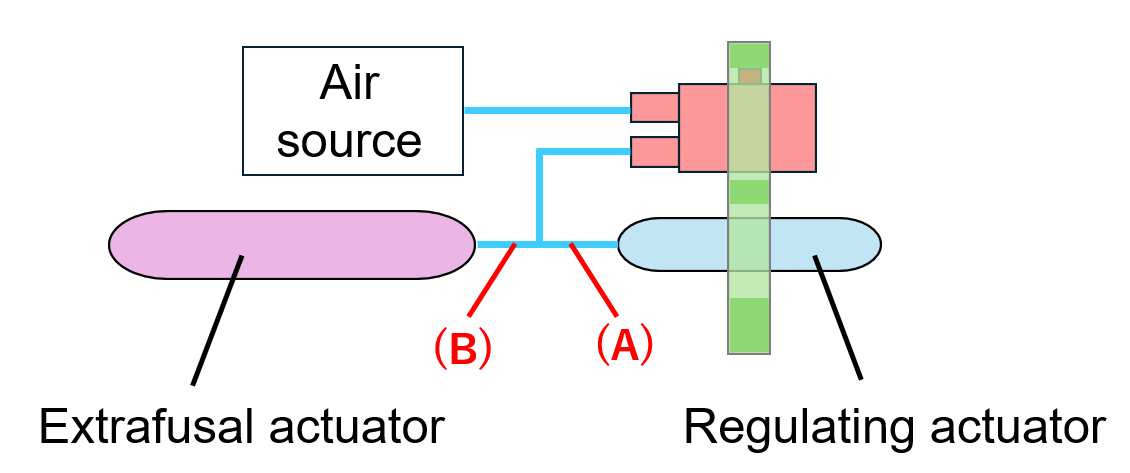}
                  \caption{Experimental setup for identifying the dominant flow restriction paths to suppress oscillations.}
                  \label{Fig:WhichFlowPath}
                \end{figure}

                In this section, we identified the flow path that contributes to oscillation suppression. 
                The two candidate flow paths are shown in Fig.~\ref{Fig:WhichFlowPath}. 
                These paths are located (A) immediately before the regulating actuator and (B) immediately before the extrafusal actuator. 
                Although the flow path on the output side of the mechanical valve could also be a candidate, it is omitted because the target flow path was identified through a comparative experiment between (A) and (B). 
                In this experiment, to identify the flow path contributing to oscillation suppression, the orifice diameters for intake and exhaust in each path were restricted to be sufficiently small ($0.4\,\mathrm{mm}$, which is $16\%$ of the diameter before restriction) for comparison. 
                
\begin{figure}[h]
                    \centering
                    \subfloat[No flow restriction]{
                    \includegraphics[scale=0.2]{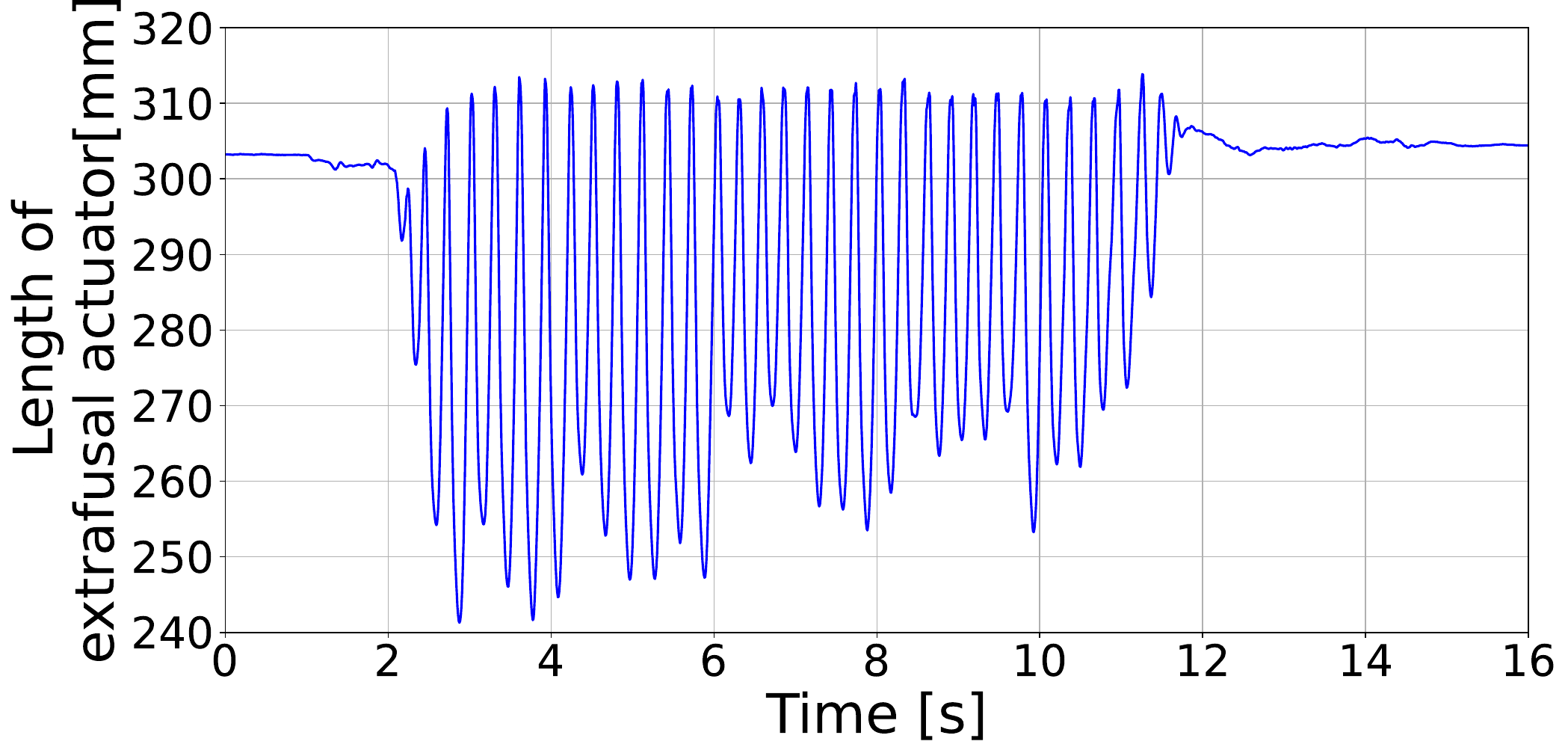}
                    \label{fig:NoLimitation_2muscle}
                    }
                    \subfloat[Restriction in (A) (0.4~mm orifice diameter)]{
                    \includegraphics[scale=0.2]{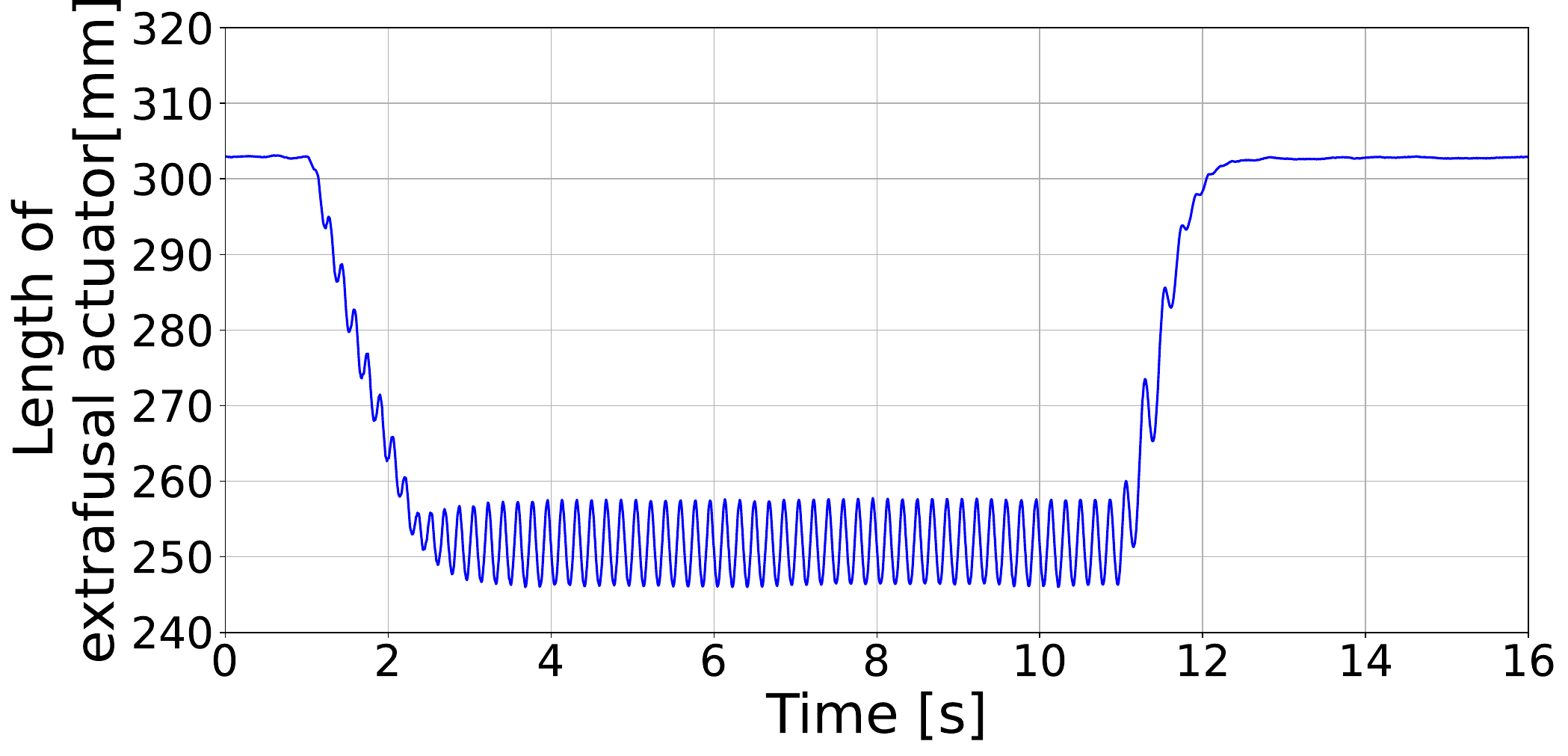}
                    \label{fig:PathA_InOut_0.4mm_2muscle}
                    }
                    \\[1em]
                    \subfloat[Restriction in (B) (0.4~mm orifice diameter)]{
                    \includegraphics[scale=0.2]{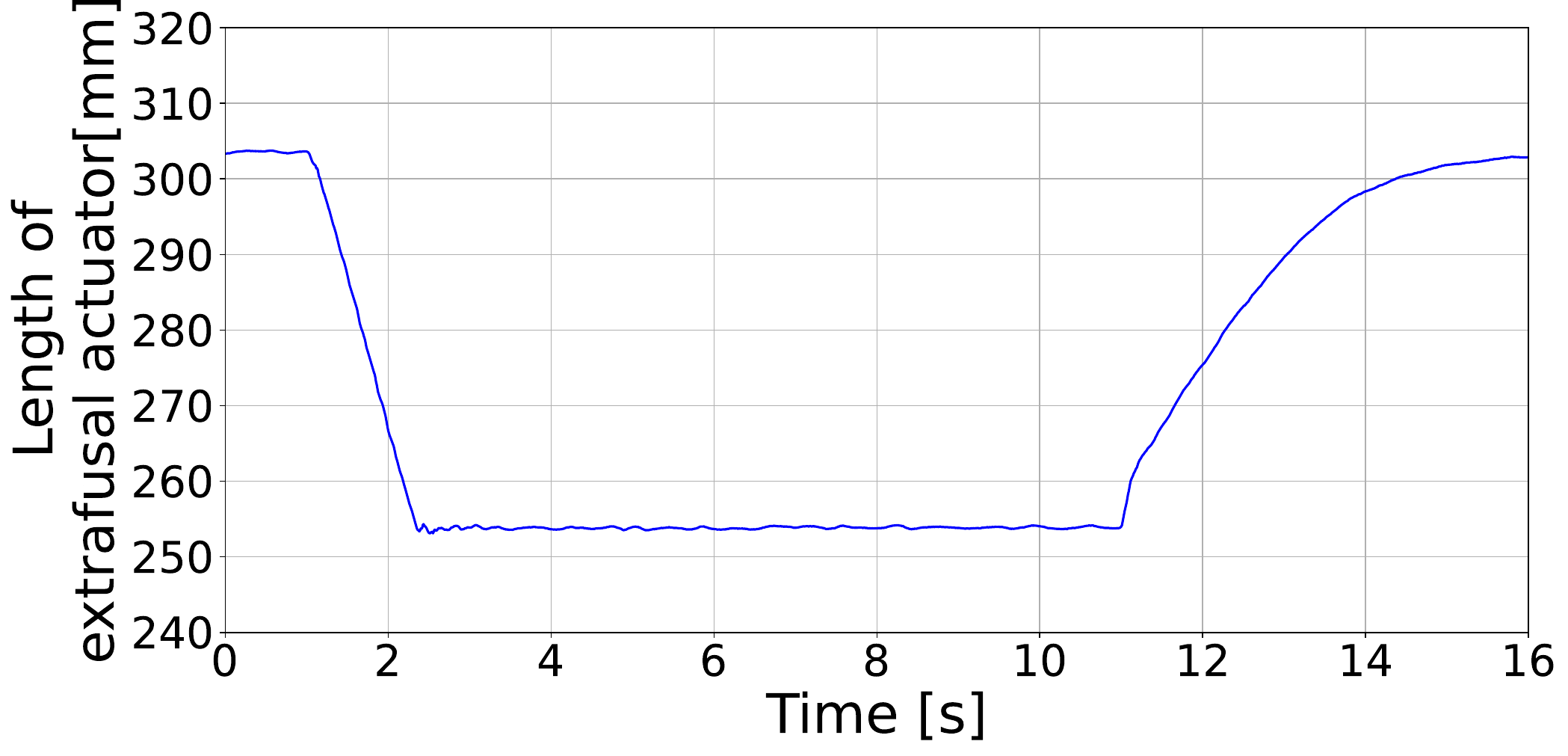}
                    \label{fig:PathB_InOut_0.4mm_2muscle}
                    }
                    \caption{Experimental results of the extrafusal actuator length showing the effects of flow restriction at different locations. The figure demonstrates that flow restriction in (B) effectively suppresses oscillations during operation, whereas oscillations persist when the flow is restricted in (A) or remains unrestricted.}
                    \label{Fig:2muscle_WhichFlowPath}
                  \end{figure}

                Fig.~\ref{Fig:2muscle_WhichFlowPath} shows the time history of the extrafusal actuator length depending on the presence or absence of flow restriction.                
                Here, Fig.~\ref{fig:NoLimitation_2muscle} shows the length change when the flow rate of each path is not restricted, Fig.~\ref{fig:PathA_InOut_0.4mm_2muscle} shows the length change when only the flow rate of (A) is restricted, and Fig.~\ref{fig:PathB_InOut_0.4mm_2muscle} shows the length change when only the flow rate of (B) is restricted.

                As shown in Fig.~\ref{fig:PathA_InOut_0.4mm_2muscle}, when (A) was restricted, the oscillations did not subside and persisted throughout the operation time of the experimental setup ($t=1\text{--}12\,\mathrm{s}$). 
                On the other hand, as seen in Fig.~\ref{fig:PathB_InOut_0.4mm_2muscle}, restricting the flow rate in (B) successfully eliminated the oscillations during operation. 
                These results demonstrate that flow restriction in (B) (located immediately before the extrafusal actuator) is effective for oscillation suppression. 
                Accordingly, the flow rate of (B) will be restricted in the following experiments. 

            \subsubsection{Effect of Flow Restriction Direction on Oscillation Suppression}
                In this experiment, the influence of the intake flow rate B$_{\mathrm{in}}$ and the exhaust flow rate B$_{\mathrm{out}}$ through the flow path at (B) was investigated. 
                In this system, the extrafusal actuator intakes air (B$_{\mathrm{in}}$) when the actuator contracts and exhausts air (B$_{\mathrm{out}}$) when it extends. 
                Therefore, it is expected that flow restriction of B$_{\mathrm{in}}$ is effective for suppressing oscillations during contraction, while restriction of B$_{\mathrm{out}}$ is effective for eliminating oscillations during extension. 
                In this experiment, we investigated the relationship between the direction of flow restriction in (B) and the oscillation phenomenon by setting the inner diameter of one of the two flow directions to be sufficiently small ($0.4$~mm, which is $16\%$ of the diameter before restriction) for comparison. 
                
                  \begin{figure}[h]
                    \centering
                    \subfloat[B$_{\mathrm{in}}$: $0.4$~mm orifice diameter, B$_{\mathrm{out}}$: unrestricted]{
                    \includegraphics[scale=0.2]{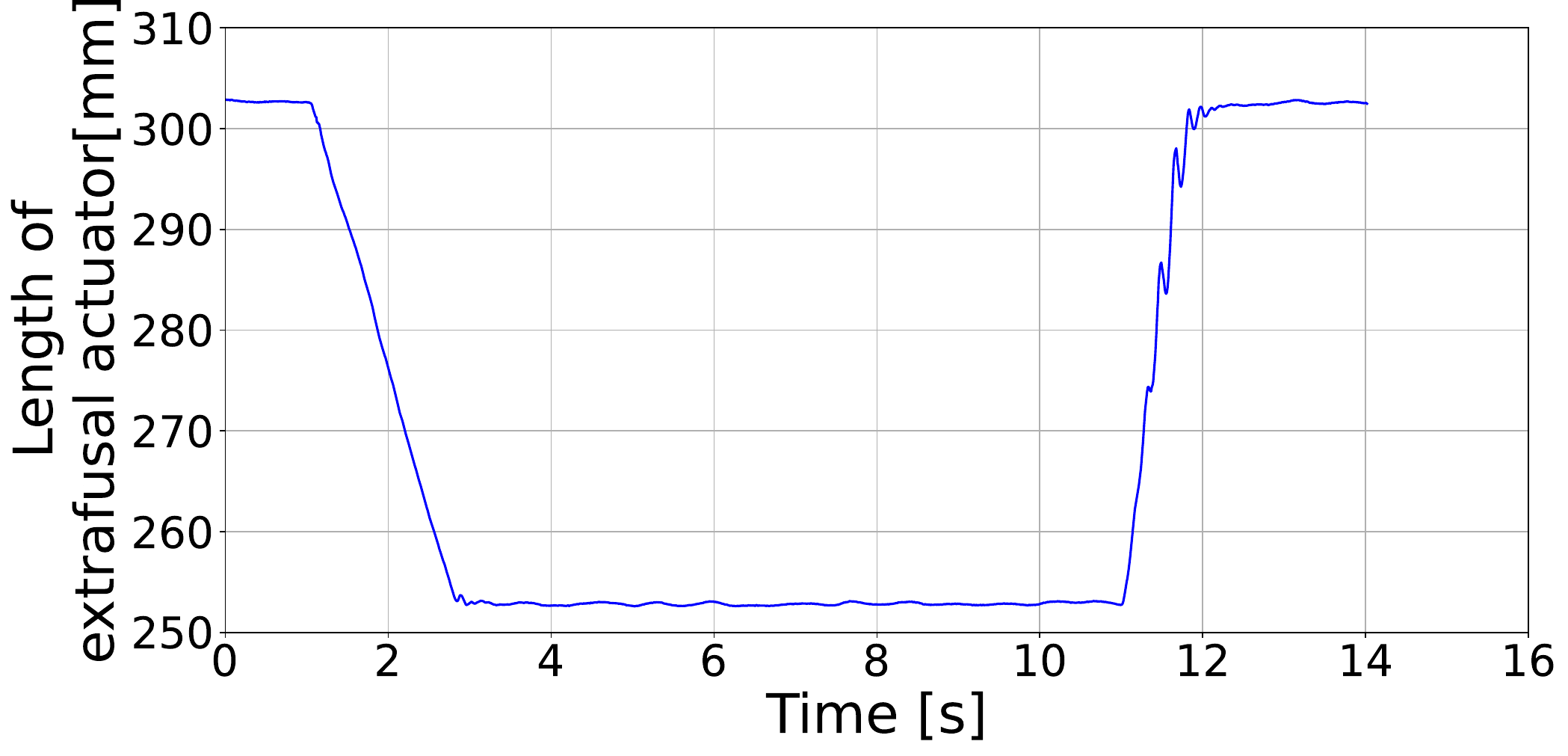}
                    \label{fig:in_0.4mm/out_NoLimitation}
                    }
                    \subfloat[B$_{\mathrm{in}}$: unrestricted , B$_{\mathrm{out}}$: $0.4$~mm orifice diameter]{
                    \includegraphics[scale=0.2]{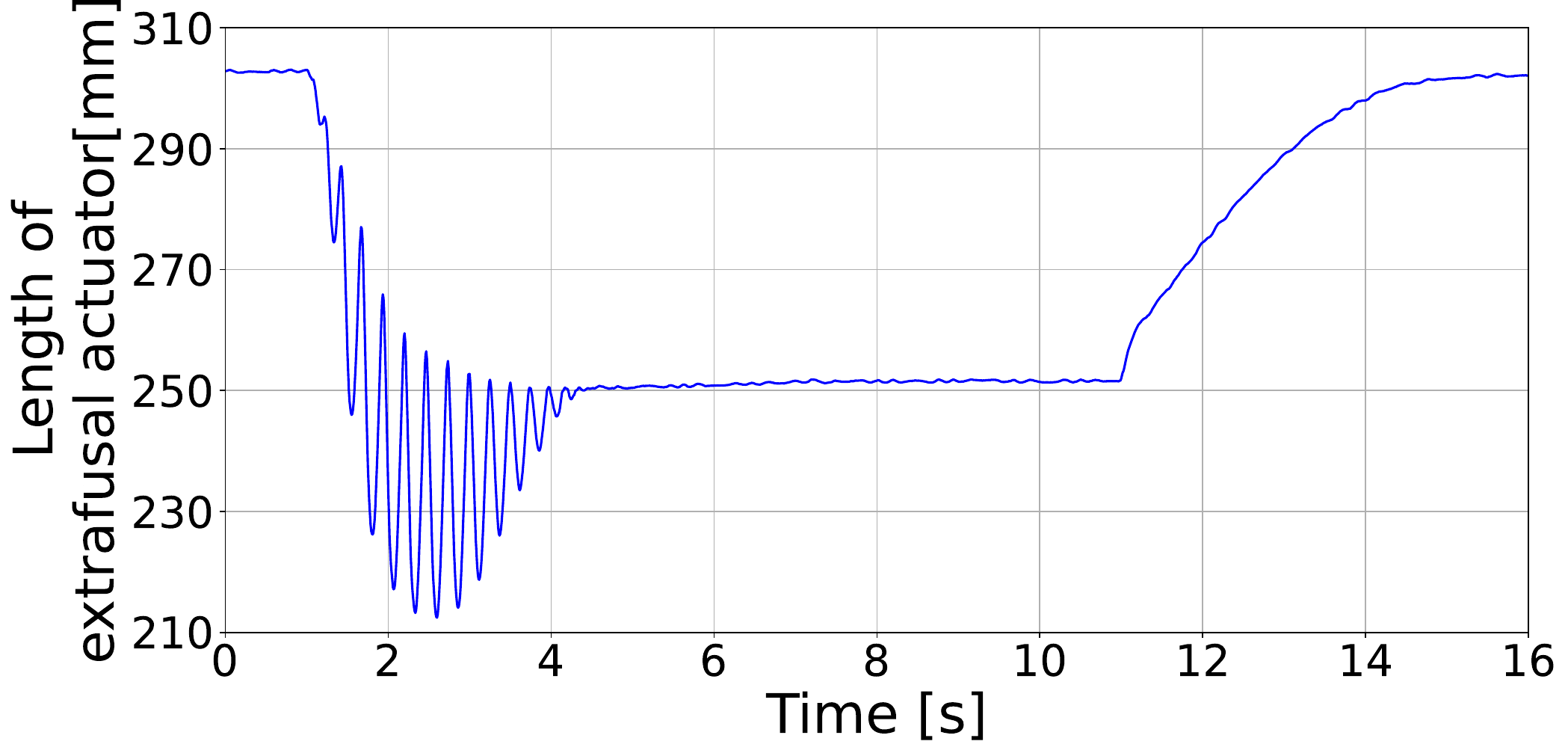}
                    \label{fig:in_NoLimitation/out_0.4mm}
                    }
                    \caption{Experimental results of the extrafusal actuator length comparing the effects of intake and exhaust flow restrictions in (B). The figure shows that restricting the intake flow rate (B$_{\mathrm{in}}$) eliminates oscillations during contraction, whereas restricting the exhaust flow rate (B$_{\mathrm{out}}$) is effective for suppressing oscillations during extension.}
                    \label{Fig:2muscle_ExActu_Direction}
                  \end{figure}   
                  
                The experimental results are shown in Fig.~\ref{Fig:2muscle_ExActu_Direction}. 
                From Fig.~\ref{fig:in_0.4mm/out_NoLimitation}, when only the intake direction B$_{\mathrm{in}}$ was restricted, the oscillations during contraction (around $t = 1\text{--}3\,\mathrm{s}$) were eliminated, but the oscillations during extension (after $t = 11\,\mathrm{s}$) slightly remained. 
                On the other hand, as seen in Fig.~\ref{fig:in_NoLimitation/out_0.4mm}, when only the flow rate in the exhaust direction B$_{\mathrm{out}}$ was restricted, oscillations remained during and immediately after contraction (around $t = 1\text{--}2\,\mathrm{s}$), while those during extension (after $t = 11\,\mathrm{s}$) were eliminated.   

            \subsubsection{Investigation of Flow Restriction Parameters for Oscillation Suppression}
                In this experiment, we identified the appropriate orifice diameters for both the intake and exhaust flow paths in order to suppress oscillations while maintaining the desired flow rate.
                In this experiment, the presence of oscillations was compared by varying the orifice diameter in increments of $0.2\,\mathrm{mm}$. 

                The investigation of the orifice diameter was conducted in two stages, B$_{\mathrm{in}}$ and B$_{\mathrm{out}}$. 
                First, the conditions for suppressing oscillations during contraction were identified by varying the orifice diameter of B$_{\mathrm{in}}$. 
                Next, after fixing B$_{\mathrm{in}}$ at the aforementioned condition, the conditions for suppressing oscillations during extension were identified by varying the orifice diameter of B$_{\mathrm{out}}$. 

\begin{figure}[h]
                    \centering
                    \subfloat[B$_{\mathrm{in}}$: $0.8\,\mathrm{mm}$ orifice diameter , B$_{\mathrm{out}}$: unrestricted]{
                    \includegraphics[scale=0.22]{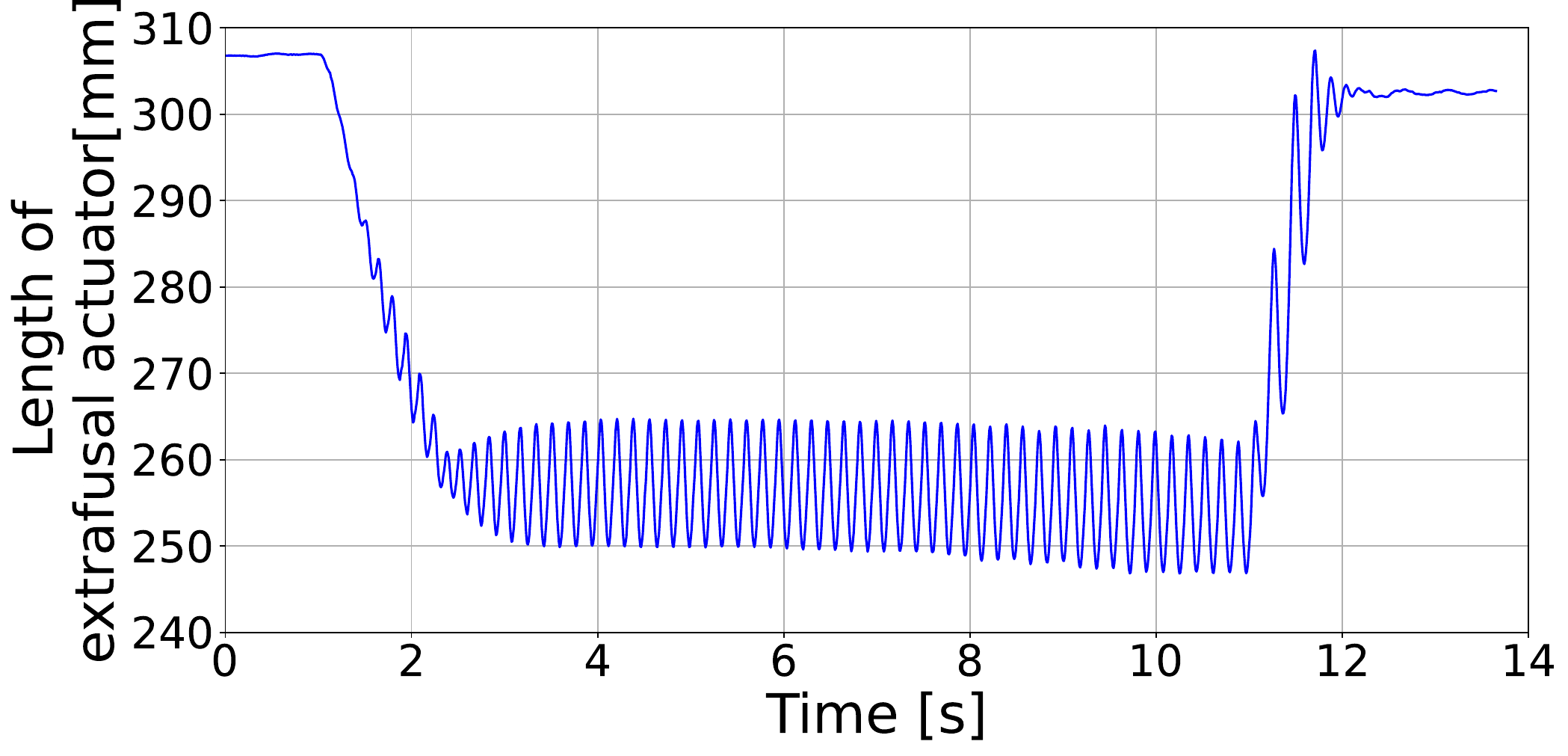}
                    \label{fig:In_0.8/Out_NoLimitation}
                    }
                    \subfloat[B$_{\mathrm{in}}$: $0.6\,\mathrm{mm}$ orifice diameter , B$_{\mathrm{out}}$: unrestricted]{
                    \includegraphics[scale=0.22]{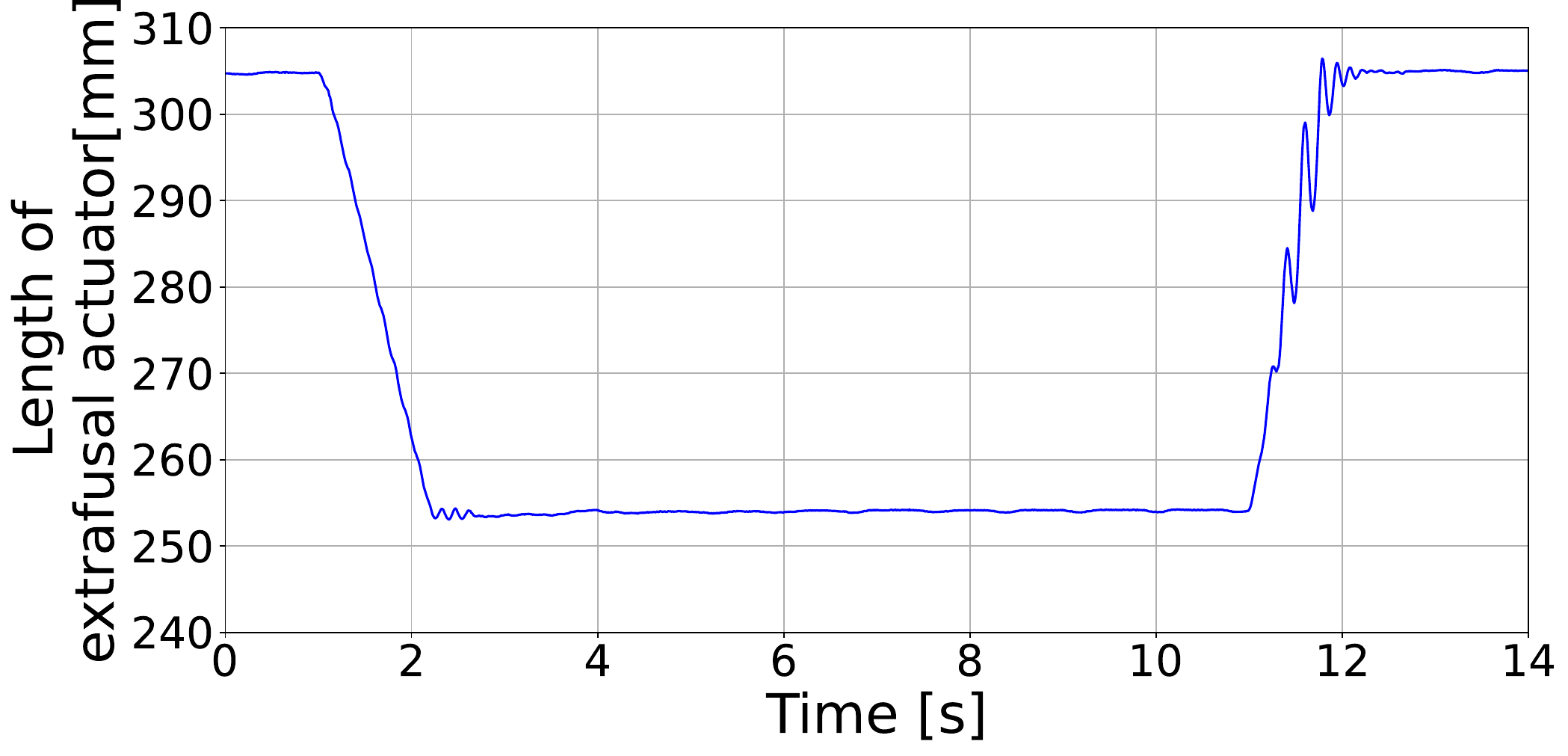}
                    \label{fig:In_0.6/Out_NoLimitation}
                    }
                    \caption{Experimental results of the extrafusal actuator length under various orifice diameters in the intake flow path (B$_{\mathrm{in}}$). The figure indicates that an orifice diameter of $0.6\,\mathrm{mm}$ effectively suppresses oscillations both during the contraction phase and while the actuator is contracted, identifying the optimal parameter for stable intake flow. }
                    \label{Fig:2muscle_Oscilation_BestPara_Bin}
                  \end{figure}

\begin{figure}[h]
                    \centering
                    \subfloat[B$_{\mathrm{in}}$: $0.6\,\mathrm{mm}$ orifice diameter , B$_{\mathrm{out}}$: $1.0\,\mathrm{mm}$ orifice diameter]{
                    \includegraphics[scale=0.22]{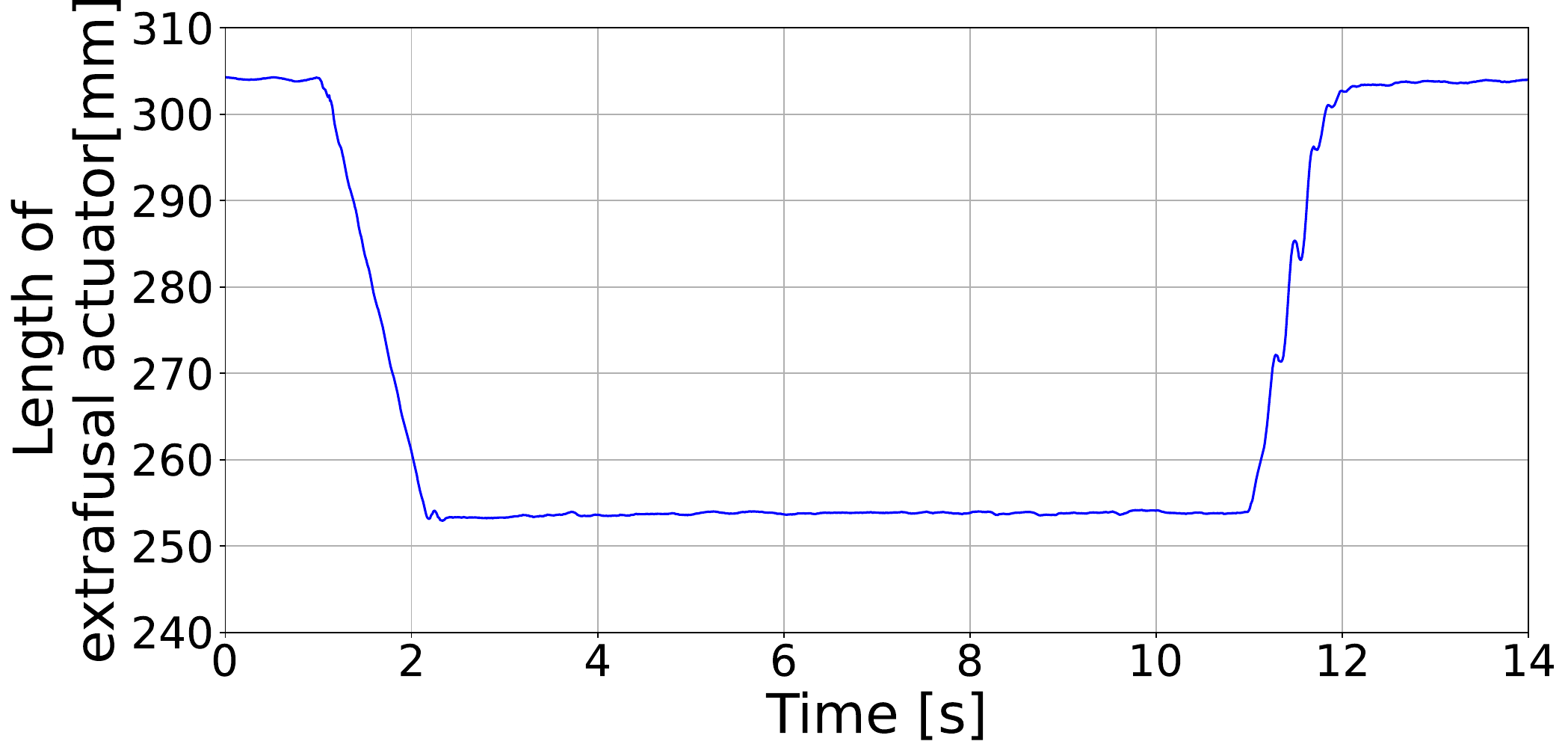}
                    \label{fig:In_0.8/Out_1.0}
                    }
                    \subfloat[B$_{\mathrm{in}}$: $0.6\,\mathrm{mm}$ orifice diameter , B$_{\mathrm{out}}$: $0.8\,\mathrm{mm}$ orifice diameter]{
                    \includegraphics[scale=0.22]{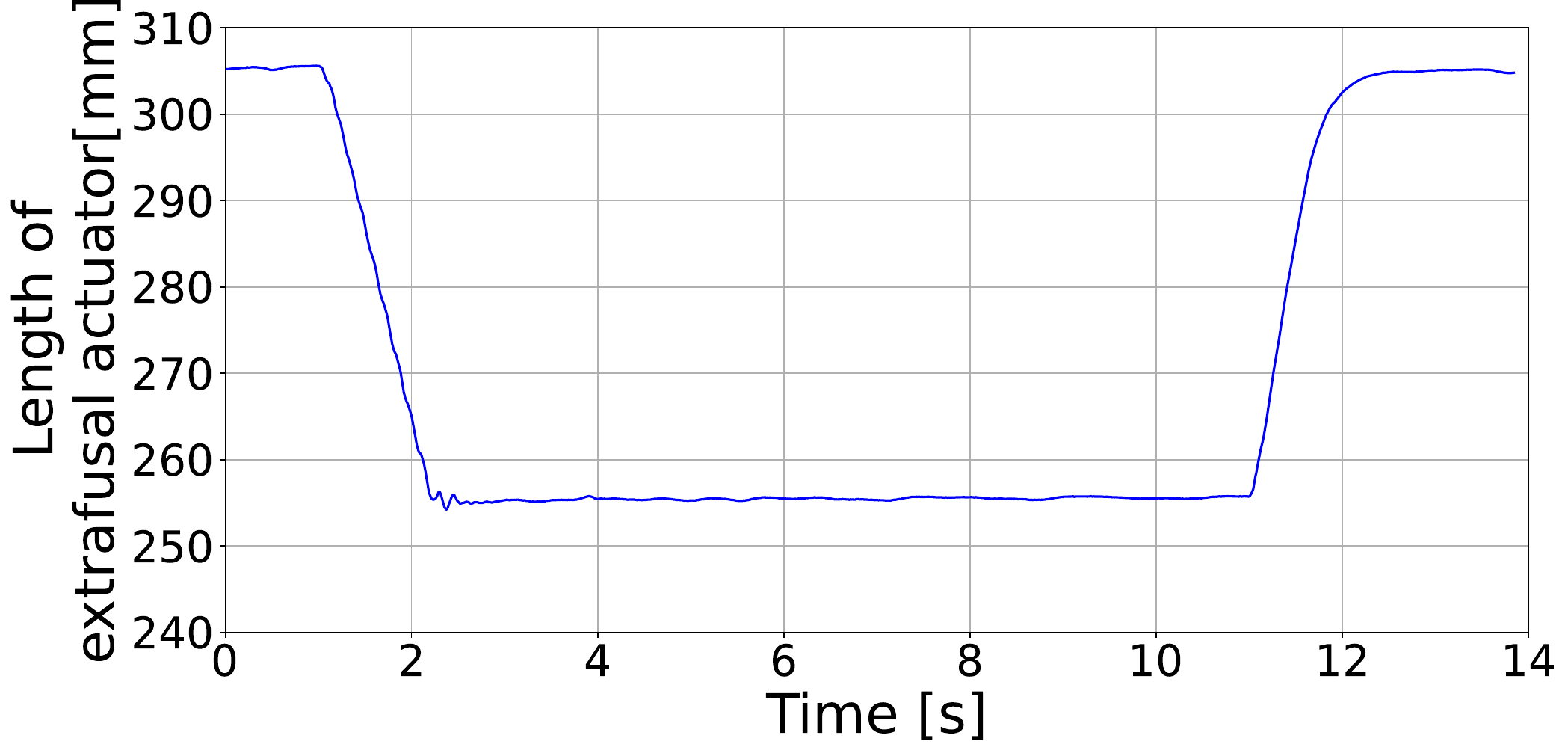}
                    \label{fig:In_0.6/Out_0.8}
                    }
                    \caption{Experimental results of the extrafusal actuator length under different orifice diameters in the exhaust flow path (B$_{\mathrm{out}}$) with B$_{\mathrm{in}}$ fixed at $0.6\,\mathrm{mm}$. The figure shows that an orifice diameter of $0.8\,\mathrm{mm}$ eliminates oscillations during operation, identifying the optimal parameter for stable exhaust flow when combined with the $0.6\,\mathrm{mm}$ intake restriction. }
                    \label{Fig:2muscle_Oscilation_BestPara_Bout}
                  \end{figure}

                The experimental results are shown in Fig.~\ref{Fig:2muscle_Oscilation_BestPara_Bin} and Fig.~\ref{Fig:2muscle_Oscilation_BestPara_Bout}. 
                Fig.~\ref{Fig:2muscle_Oscilation_BestPara_Bin} shows the comparative experiments when the flow restriction of B$_{\mathrm{in}}$ was varied. 
                From Fig.~\ref{fig:In_0.6/Out_NoLimitation}, it can be seen that an orifice diameter of $0.6\,\mathrm{mm}$ ($24\%$ of the original internal diameter) in B$_{\mathrm{in}}$ suppresses oscillations both during contraction (around $t = 1\text{--}2\,\mathrm{s}$) and while contracted (around $t = 2\text{--}11\mathrm{s}$). 
                Next, Fig.~\ref{Fig:2muscle_Oscilation_BestPara_Bout} shows the comparative experiments where the flow restriction of B$_{\mathrm{out}}$ was varied while the orifice diameter of B$_{\mathrm{in}}$ was fixed at $0.6\,\mathrm{mm}$. 
                From Fig.~\ref{fig:In_0.6/Out_0.8}, it is observed that an orifice diameter of $0.8\,\mathrm{mm}$ ($32\%$ of the original internal diameter) in B$_{\mathrm{out}}$ eliminates oscillations during operation. 
                In the following experiments, the orifice diameters of B$_{\mathrm{in}}$ and B$_{\mathrm{out}}$ were fixed at $0.6\,\mathrm{mm}$ and $0.8\,\mathrm{mm}$, respectively.
                
    \subsection{Performance Evaluation of the Proposed Actuator System}
        \subsubsection{Evaluation of Length Control Performance}
            In this experiment, we investigated the length control performance of the PAMs using the proposed method. 
            
            In this experiment, four sinusoidal inputs with different frequencies were applied to the servo motor for comparison. 
            The amplitude of the sinusoidal input to the servo motor was 400 deg; at this time, the wire winding displacement varied between $0\,\mathrm{mm}$ and $52\,\mathrm{mm}$. 
            The tracking performance of the extrafusal actuator length relative to the input displacement was evaluated using the root mean square error (RMSE\footnote{$\text{RMSE} = \sqrt{\frac{1}{n} \sum_{i=1}^{n} (d_{\mathrm{target},i} - d_{\mathrm{measured},i})^2}$, where $d_{\mathrm{target},i}$ and $d_{\mathrm{measured},i}$ are the target and measured displacements of the extrafusal actuator at the $i$-th sample, respectively, and $n$ is the number of samples.}).
            To eliminate the influence of slack in the intrafusal actuator, the experiment was initiated with the servo motor rotated by a certain amount to ensure tension was maintained in the wire. 

\begin{figure}[t]
                    \centering
                    \subfloat[$0.15\,\mathrm{Hz}$ (RMSE: 0.74~mm)]{
                    \includegraphics[scale=0.19]{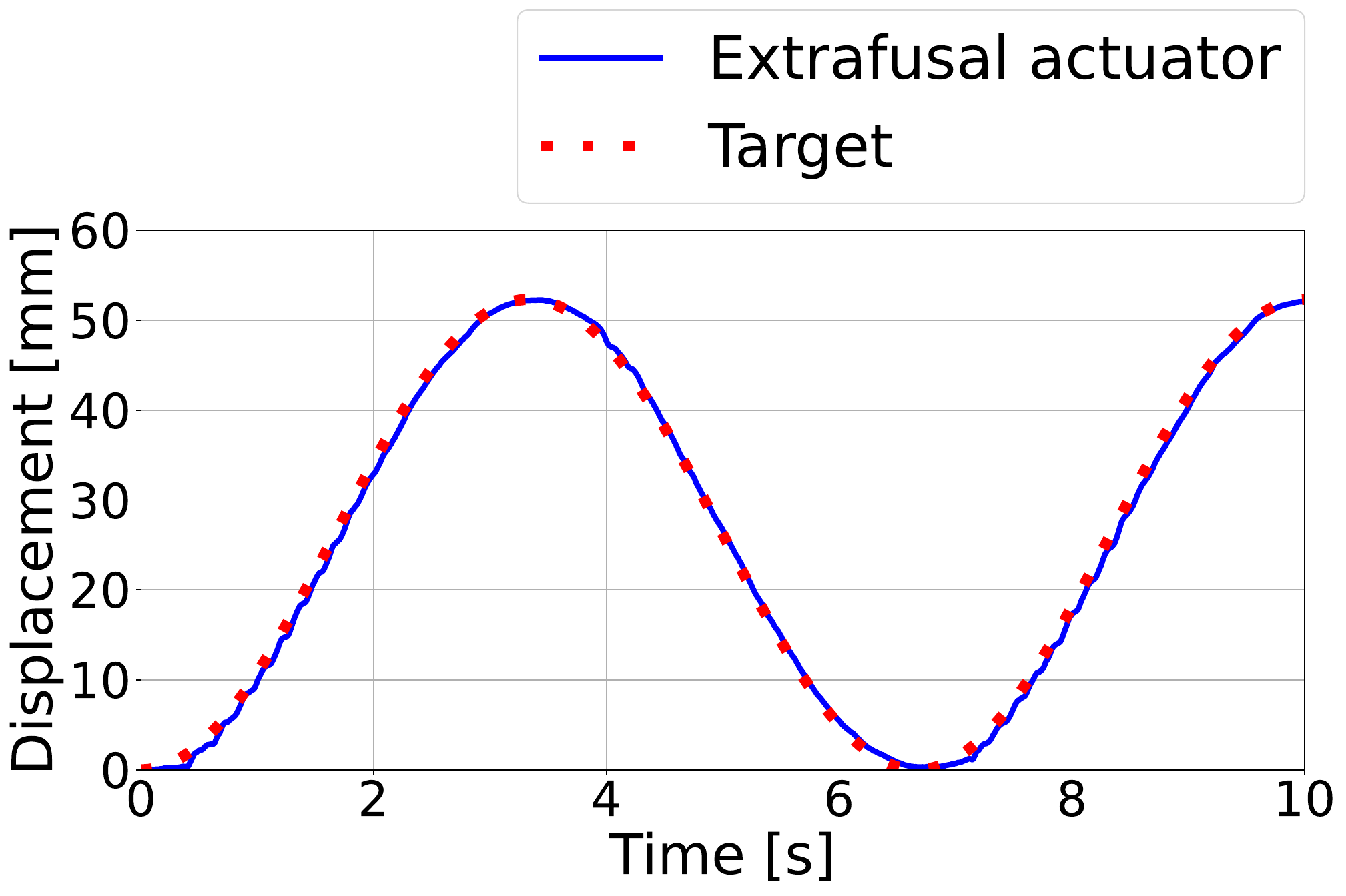}
                    \label{fig:0.15Hz}
                    }
                    \subfloat[$0.20\,\mathrm{Hz}$ (RMSE: 1.05~mm)]{
                    \includegraphics[scale=0.19]{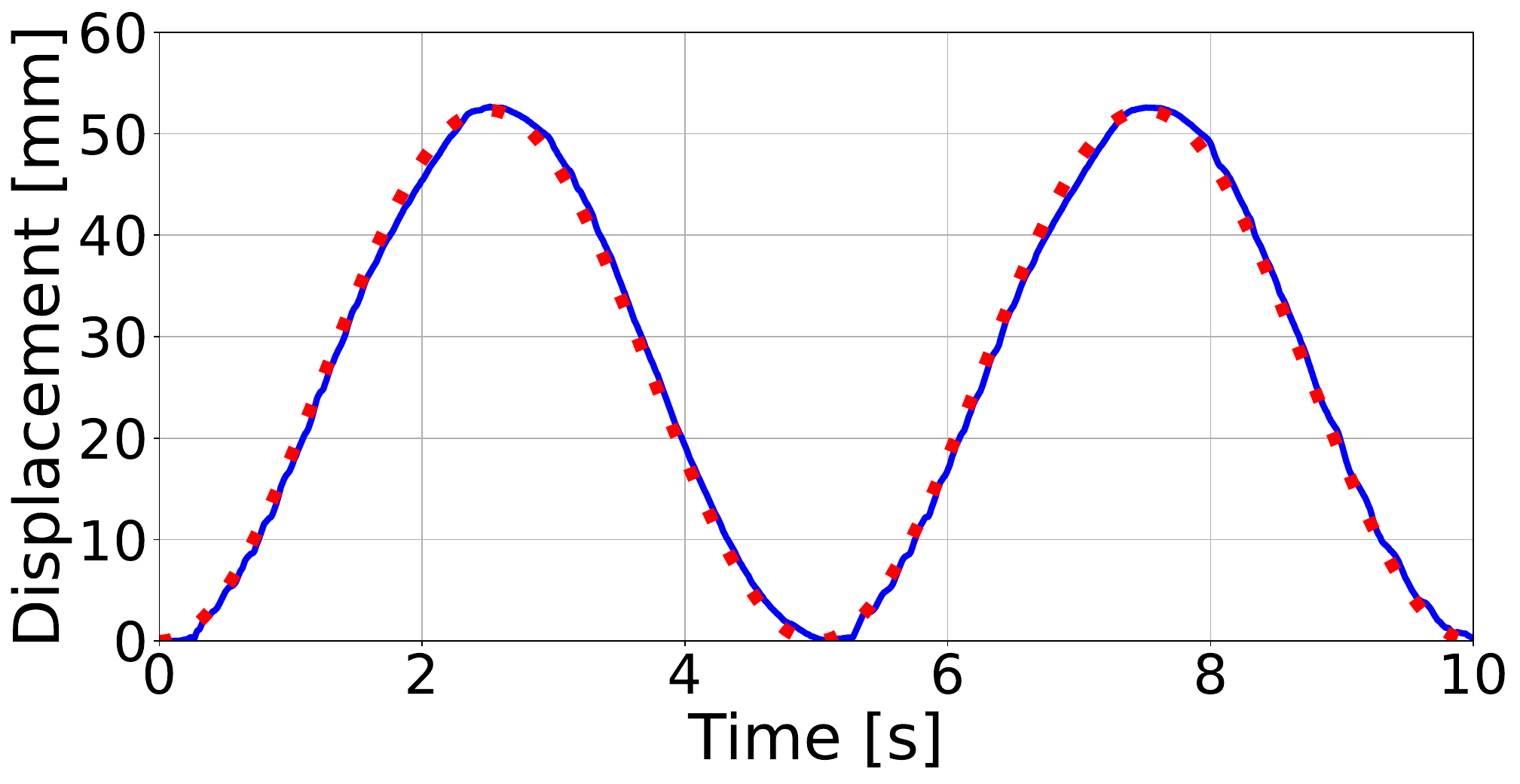}
                    \label{fig:0.20Hz}
                    }
                    \\[1em]
                    \subfloat[$0.30\,\mathrm{Hz}$ (RMSE: 3.65~mm)]{
                    \includegraphics[scale=0.19]{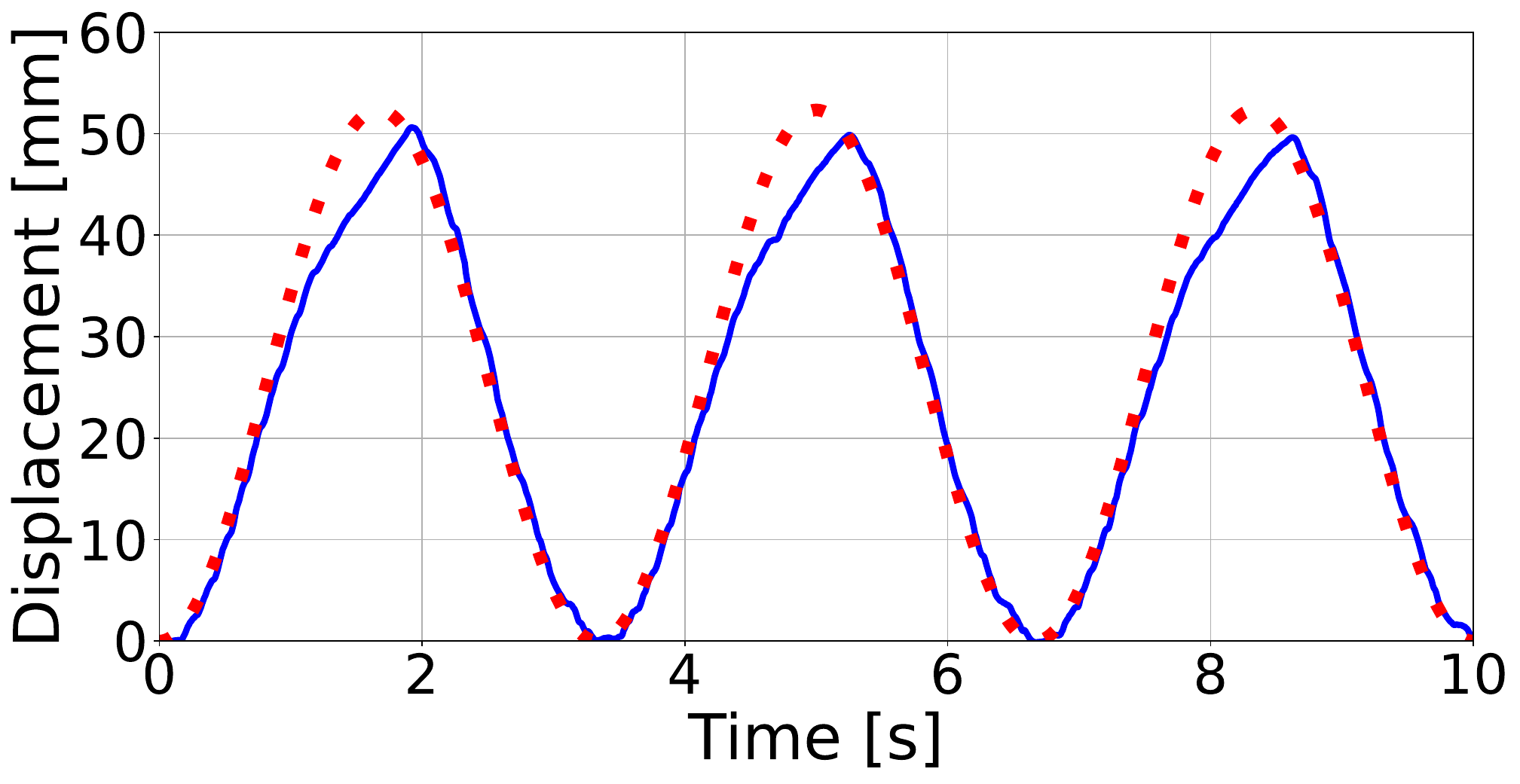}
                    \label{fig:0.30Hz}
                    }
                    \subfloat[$0.40\,\mathrm{Hz}$ (RMSE: 6.98~mm)]{
                    \includegraphics[scale=0.19]{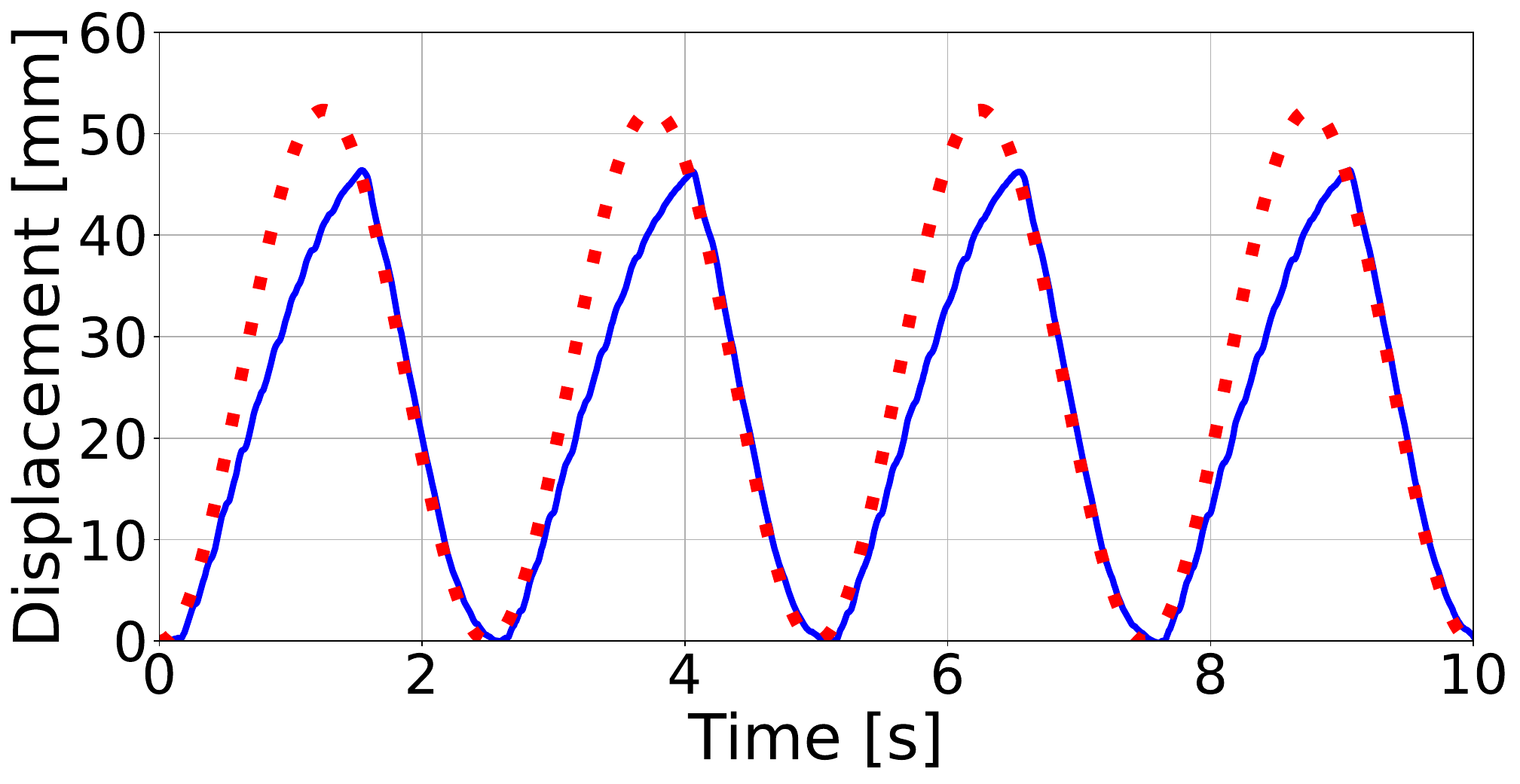}
                    \label{fig:0.40Hz}
                    }                    
                    \caption{The tracking performance relative to the target command, evaluated using RMSE. The figure demonstrates high-precision tracking at frequencies of $0.20\,\mathrm{Hz}$ or lower, with the RMSE remaining within $1.05\,\mathrm{mm}$ (error $\le 2.01\%$). At input frequencies of $0.30\,\mathrm{Hz}$ or higher, tracking performance degrades due to delays in the pneumatic system}
                    \label{Fig:2muscle_ControlPerfomance}
                \end{figure}

            Fig.~\ref{Fig:2muscle_ControlPerfomance} shows the time histories of the extrafusal actuator length. 
            Note that the vertical axis of each graph represents the displacement of the extrafusal and intrafusal actuators (wire) relative to their initial lengths at the start of the experiment. 
            The solid blue lines in the figure represent the displacement of the extrafusal actuator, and the dotted red lines represent the wire winding displacement derived from the servo motor input command value and the pulley diameter. 
            Fig.~\ref{fig:0.15Hz} and Fig.~\ref{fig:0.20Hz} show that at input frequencies of $0.20\,\mathrm{Hz}$ or lower, the RMSE remains within $1.05\,\mathrm{mm}$ (error $2.01\%$\footnote{The error defined as $\mathrm{RMSE}/52$, where $52\,\mathrm{mm}$ is the maximum input displacement}), confirming that the extrafusal actuator displacement follows the input command with high precision. 
            Compared to previous studies \cite{Dao2021FPID,shin2021detailed} that investigated length control of PAMs and evaluated tracking errors using RMSE (errors ranging from $2.90\text{--}4.17\,\%$), the proposed method improved the normalized RMSE by at least $0.89$ percentage points\footnote{Since previous studies reported only RMSE values, we normalized the values based on the given experimental conditions.}.

            On the other hand, Fig.~\ref{fig:0.30Hz} and Fig.~\ref{fig:0.40Hz} reveal that when the input frequency exceeds $0.30$~Hz, the tracking performance of the extrafusal actuator relative to the input wire winding displacement (red dotted line) decreases. 
            The tracking performance deteriorates during contraction, with the RMSE reaching $3.65\,\mathrm{mm}$ (error $6.98\%$) at $0.30\,\mathrm{Hz}$ and $6.98\,\mathrm{mm}$ (error $13.35\%$) at $0.40\,\mathrm{Hz}$. 
            This is presumed to be because the extrafusal actuator could not generate sufficient tension due to delays in the pneumatic system, thereby hindering the contraction of the intrafusal actuator.    

        \subsubsection{Evaluation of Load Compensation Performance}
        \label{LoadCompensationPerformance}
            In this section, we investigated the length compensation performance when the external load is increased. 

            In the experiment, we conducted a load variation test by changing the number of suspended weight plates while the extrafusal actuator remained stationary at a constant length. 
            Before the experiment began, the target winding displacement of the wire in the intrafusal actuator was set to $52\,\mathrm{mm}$, and a $2.5\,\mathrm{kg}$ metal dumbbell shaft was suspended as an initial load. 
            After the experiment commenced, $2.5\,\mathrm{kg}$ weight plates were added stepwise three times, followed by two additions of $5.0\,\mathrm{kg}$ weight plates. 
            The final total load reached $20.0\,\mathrm{kg}$. 
            Note that an interval of approximately 20 seconds was provided after each weight plate was added. 
            
\begin{figure}[h]
                  \centering
                  \includegraphics[scale=0.25]{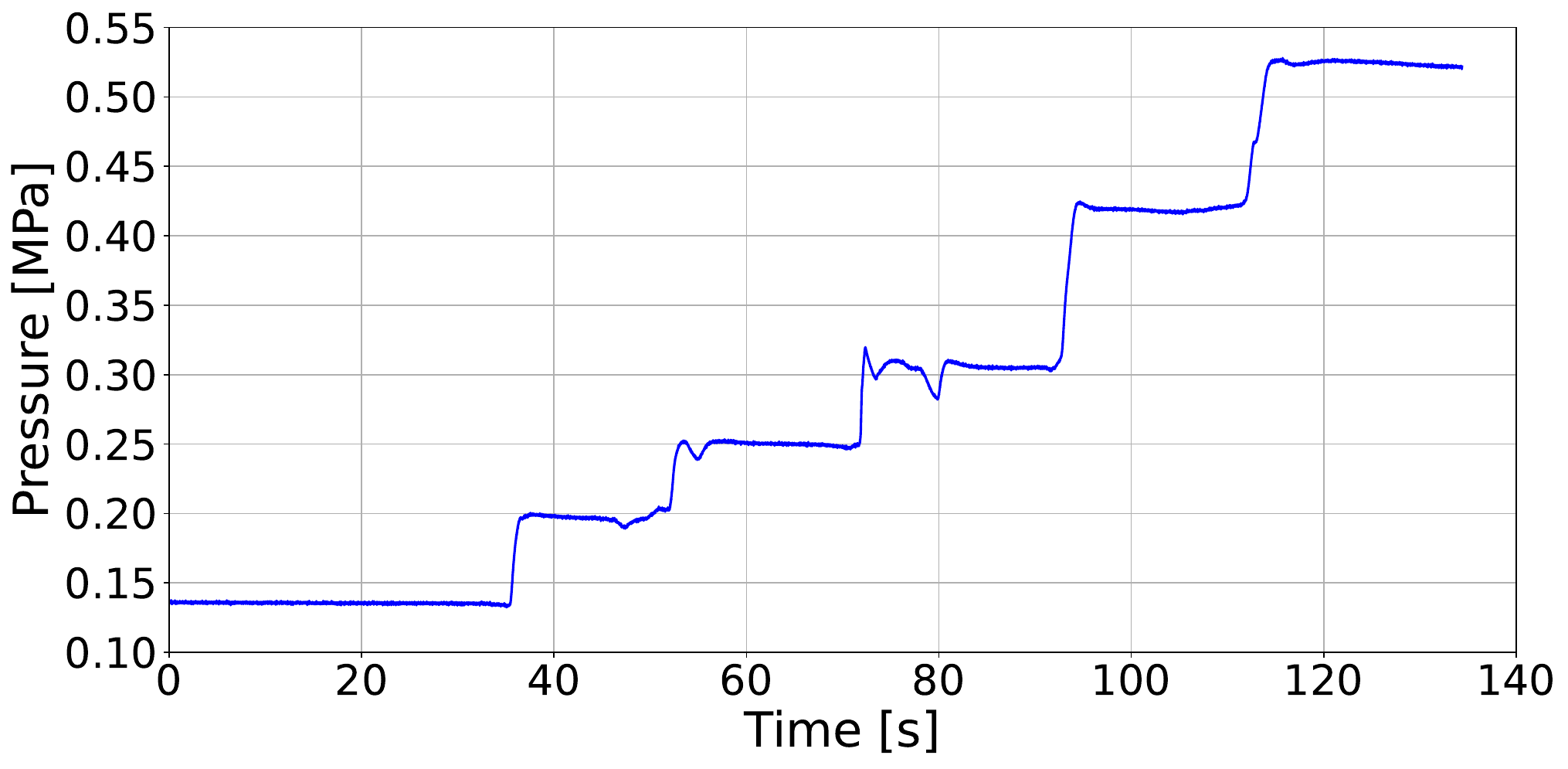}
                  \caption{Experimental results of the internal pressure in the extrafusal actuator during the load variation test. The figure indicates that the mechanically implemented pneumatic feedback operates in response to the increase in load, causing the internal pressure to rise in a staircase pattern from $0.14\,\mathrm{MPa}$ to $0.52\,\mathrm{MPa}$ as the total load reaches $20.0\,\mathrm{kg}$.}
                  \label{Fig:Compensation_Pressure}
                \end{figure}

\begin{figure}[h]
                  \centering
                  \includegraphics[scale=0.25]{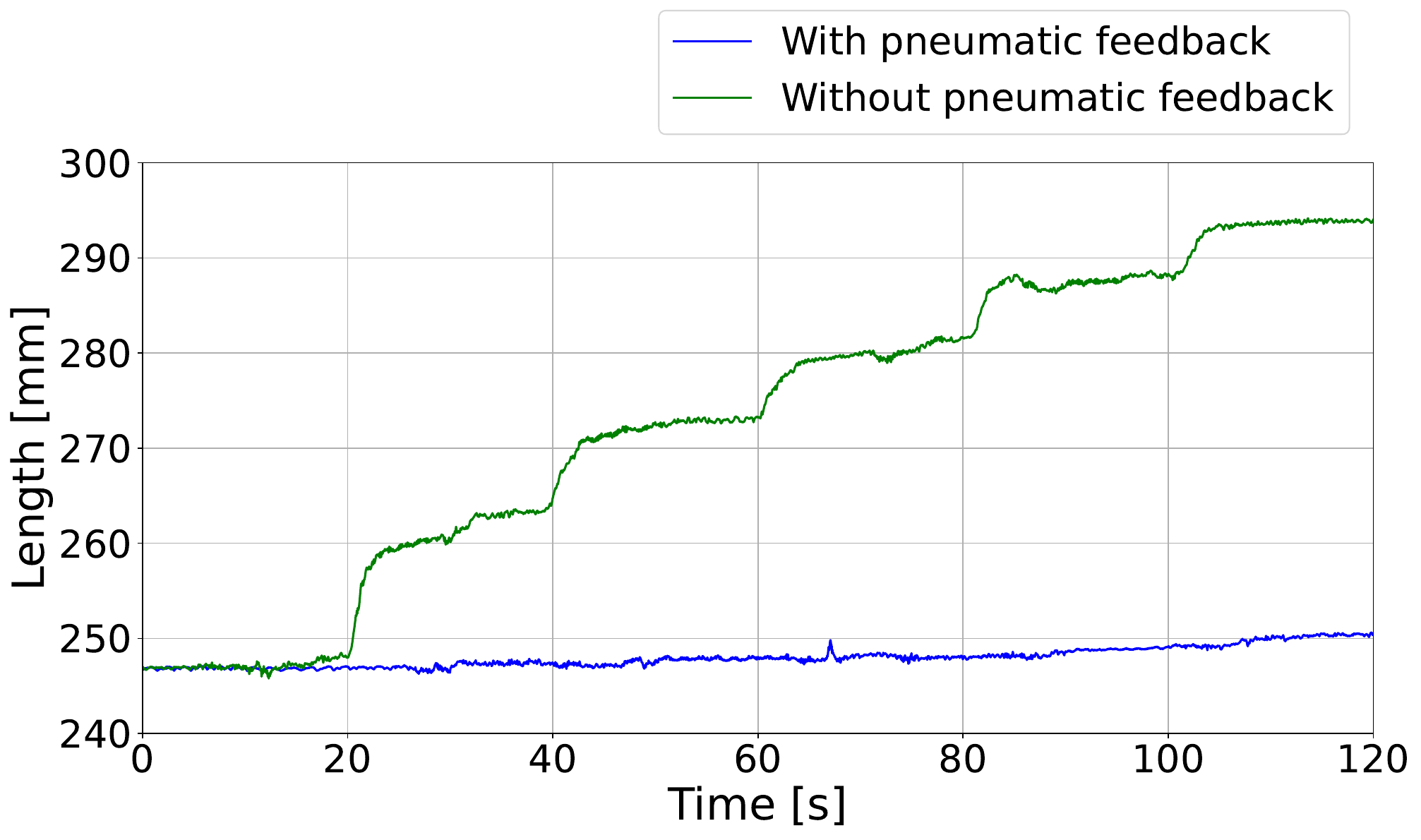}
                  \caption{Load compensation performance evaluated by comparing the length change of the extrafusal actuator with and without the proposed pneumatic feedback. The figure demonstrates that the proposed method suppresses length displacement to $3.8\,\mathrm{mm}$ ($1.5\%$ of the pre-load-increase length), compared to $46.7\,\mathrm{mm}$ ($18.8\%$) without feedback, thereby confirming high robustness against external load variations.}
                  \label{Fig:Compensation_Length}
                \end{figure}

            The experimental results are shown in Fig.~\ref{Fig:Compensation_Pressure} and Fig.~\ref{Fig:Compensation_Length}. First, Fig.~\ref{Fig:Compensation_Pressure} shows the time history of the internal pressure of the extrafusal actuator when weights are added stepwise. 
            The figure indicates that the mechanically implemented pneumatic feedback operates in response to the increase in load, and the internal pressure of the extrafusal actuator rises in a staircase pattern from $0.14\,\mathrm{MPa}$ to $0.52\,\mathrm{MPa}$.          

            Next, Fig.~\ref{Fig:Compensation_Length} shows the time history of the extrafusal actuator length in the same experiment. 
            The blue line in the figure represents the actuator length with the proposed method, while the green line represents the length when a constant pressure ($0.14\,\mathrm{MPa}$) is applied to the extrafusal actuator. 
            As a result of the experiment, the length change without pneumatic feedback was $46.7\,\mathrm{mm}$ (a change rate of $18.8\%$ relative to the start of the experiment), whereas the length change with pneumatic feedback was limited to $3.8\,\mathrm{mm}$ (1.5$\%$). 
            These results demonstrate that the proposed method possesses high load compensation performance against varying external loads.  

\section{CONCLUSIONS}
    In this study, we proposed an actuator system (InExSA) that reproduces the interactions between intrafusal and extrafusal muscles in an engineering implementation, and conducted its performance evaluation. 

    The main findings of this study are the following three points.
    First, for the oscillation phenomena that pose challenges in pneumatic control systems, we identified the flow paths contributing to oscillation suppression and determined the effective orifice diameters and their directions. 
    Second, length control experiments clarified the effectiveness and current limitations of the proposed system. In the frequency range of $0.20\,\mathrm{Hz}$ or lower, the tracking error (RMSE) relative to the target command was suppressed to $1.05\,\mathrm{mm}$ or less. 
    Conversely, length control performance degraded as the input frequency increased. 
    Notably, within the bandwidth where tracking performance is maintained, the displacement of the intrafusal wire remains synchronized with that of the PAMs. This offers the advantage that the PAM length can be geometrically estimated from the servo motor angle without requiring additional state estimators.
    Third, load compensation experiments clarified the effectiveness of the proposed system. 
    This actuator system autonomously adjusted the air pressure for loads up to $20.0\,\mathrm{kg}$, maintaining the length error of the PAMs within $3.8\,\mathrm{mm}$ ($1.5\%$ of the length before loading). 
    Furthermore, while the supply pressure from the pneumatic source was $0.75\,\mathrm{MPa}$, the final air pressure in this experiment remained below $0.55\,\mathrm{MPa}$. 
    This remaining pressure margin may indicate that the proposed system could compensate for loads exceeding $20.0$~kg under similar experimental conditions.
    
    Since the system operates using only a single air supply line from the pneumatic source and daisy-chained wiring for the servo motors, it is easier to implement and maintain than conventional methods that require complex wiring for electro--pneumatic valves, soft sensors, and A/D converters. 
    Future work will focus on improving the system's responsiveness and realizing high-power robotic systems that feature multiple PAMs with distributed autonomous length control.

%
%




\data{The data that support the findings of this study are available from the corresponding author upon reasonable request.}

\section*{Conflict of interest}
    The authors declare that they have no known competing financial interests or personal relationships that could have appeared to influence the work reported in this paper.



\begin{thebibliography}{99}
    
        \bibitem{hosoda2006biped} K. Hosoda, T. Takuma, and A. Nakamoto, ``Design and Control of 2D Biped that can Walk and Run with Pneumatic Artificial Muscles'', Proceedings of the 6th IEEE-RAS International Conference on Humanoid Robots, pp. 284--289, 2006.
        
        \bibitem{niiyama2007} R. Niiyama, A. Nagakubo, and Y. Kuniyoshi, ``Mowgli: A bipedal jumping and landing robot with an artificial musculoskeletal system'', Proc. IEEE Int. Conf. Robot. Autom. (ICRA), pp. 2546--2551, 2007.
        
        \bibitem{yamada2011neural} Y. Yamada, S. Nishikawa, K. Shida, R. Niiyama, and Y. Kuniyoshi, ``Neural-body coupling for emergent locomotion: A musculoskeletal quadruped robot with spinobulbar model'', 2011 IEEE/RSJ International Conference on Intelligent Robots and Systems (IROS), pp. 1499--1506, 2011.
        
        \bibitem{lei2016dynamic} J. Lei, H. Yu, and T. Wang, ``Dynamic bending of bionic flexible body driven by pneumatic artificial muscles (PAMs) for spinning gait of quadruped robot'', Chinese Journal of Mechanical Engineering, vol. 29, pp. 11--20, 2016.
        
        \bibitem{tondu2000modeling} B. Tondu and P. Lopez, ``Modeling and control of McKibben artificial muscle robot actuators'', IEEE Control Systems Magazine, vol. 20, no. 2, pp. 15--38, 2000.
    
        \bibitem{Xing2010observer} K. Xing, J. Huang, Y. Wang, J. Wu, Q. Xu, and P. Shi, ``Tracking control of pneumatic artificial muscle actuators based on sliding mode and non-linear disturbance observer'', IET Control Theory \& Applications, vol. 4, no. 10, pp. 2058--2070, 2010.
    
        \bibitem{yokoyama2017mckibben} K. Yokoyama and K. Kogiso, ``PID Position Control of McKibben Pneumatic Artificial Muscle Using Only Pressure Feedback'', Proceedings of the 2018 Annual American Control Conference (ACC), pp. 3362--3367, 2018.
        
        \bibitem{hitzmann2019muscle} A. Hitzmann, S. Ikemoto, and K. Hosoda, ``Highly-integrated muscle-spindles for pneumatic artificial muscles made from conductive fabrics'', Biomimetic and Biohybrid Systems: 8th International Conference, Living Machines 2019, vol. 8, pp. 171--182, Springer, 2019.
    
        \bibitem{takahashi2023reflex} R. Takahashi, Y. Wang, J. Wang, Y. Jiang, and K. Hosoda, ``Implementation of Basic Reflex Functions on Musculoskeletal Robots Driven by Pneumatic Artificial Muscles'', IEEE Robotics and Automation Letters, vol. 8, no. 4, pp. 1920--1926, 2023.
        
        \bibitem{park2013smart} Y.-L. Park and R. J. Wood, ``Smart pneumatic artificial muscle actuator with embedded microfluidic sensing'', Proceedings of the IEEE SENSORS 2013, pp. 1--4, 2013.
        
        \bibitem{felt2016contraction} W. Felt, K. Y. Chin, and C. D. Remy, ``Contraction Sensing With Smart Braid McKibben Muscles'', IEEE/ASME Transactions on Mechatronics, vol. 21, no. 3, pp. 1201--1209, 2016.
    
        \bibitem{bennett2001review}S. Bennett, ``The past of PID controllers'', Annual reviews in control, vol. 25, pp. 43-53, 2001.   
        
        \bibitem{turnbull1959hydraulic} D. E. Turnbull, ``The response of a loaded hydraulic servomechanism'', Proceedings of the Institution of Mechanical Engineers, vol. 173, no. 9, pp. 270--295, 1959.
    
    
        
        
        \bibitem{moriguchi2018pneumatic} H. Moriguchi, O. Ichikawa, and Y. Tagawa, ``Development of a pneumatic master-slave system comprising entirely mechanical components'', Mechanical Engineering Journal, vol. 7, no. 3, pp. 19--00430, 2020.
    
        \bibitem{wang2025review} X. Wang, K. Zheng, J. Li, H. Wang, R. Zheng, and J. Liu, ``Fluid regulation strategies and valve innovations in soft robots: A review'', Robotica, pp. 1--44, 2025. 
    
        \bibitem{Choe2023SoftValve} J. K. Choe, J. Kim, H. Song, J. Bae, and J. Kim, ``A soft, self-sensing tensile valve for perceptive soft robots'', Nature Communications, vol. 14, no. 1, p. 3942, 2023.
    
        \bibitem{Stanley2024FluidicTransistor} A. A. Stanley, E. S. Roby, and S. J. Keller, ``High-speed fluidic processing circuits for dynamic control of haptic and robotic systems'', Science Advances, vol. 10, no. 14, eadl3014, 2024.
        
        
        \bibitem{masuda2022forcereceptor} Y. Masuda, R. Wakamoto, and M. Ishikawa, ``Development of electronics-free force receptor for pneumatic actuators'', Advanced Robotics, vol. 36, no. 13, pp. 670--677, 2022.
        
        \bibitem{kandel2012principles} E. R. Kandel, J. H. Schwartz, T. M. Jessell, S. A. Siegelbaum, and A. J. Hudspeth, ``Principles of Neural Science, Fifth Edition'', McGraw-Hill, 2012.
        
        
        
        
        \bibitem{blum2017force} K. P. Blum, B. Lamotte D’Incamps, D. Zytnicki, and L. H. Ting, ``Force encoding in muscle spindles during stretch of passive muscle'', PLoS Computational Biology, vol. 13, no. 9, p. e1005767, 2017.
        
        \bibitem{masuda2025robomech} Y. Masuda, M. Ishikane, and M. Ishikawa, ``Muscle length control method for nonlinear musculoskeletal robots by mimicking intrafusal muscle fibers, Part 2: Brainless muscle length control of McKibben-type artificial muscles'', Proc. JSME Conf. Robotics and Mechatronics 2025 (ROBOMECH 2025), 2P1-H10, 2025 (in Japanese).
        
        
        
    
        \bibitem{Dao2021FPID} Q. T. Dao, D.-H. Mai, and D.-K. Nguyen, ``Experimental Validation for Pneumatic Artificial Muscles using Fuzzy PID Control'', Journal of Measurement, Control and Automation, vol. 2, no. 1, 2021.
        
        \bibitem{shin2021detailed} T. Shin, T. Ibayashi, and K. Kogiso, ``Detailed dynamic model of antagonistic PAM system and its experimental validation: Sensorless angle and torque control with UKF'', IEEE/ASME Transactions on Mechatronics, vol. 27, no. 3, pp. 1715--1726, 2021.
    
    \end{thebibliography}
\end{document}